\documentclass[lettersize,journal]{IEEEtran}
\usepackage{amsmath,amsfonts}
\usepackage{algorithmic}
\usepackage{array}
\usepackage{textcomp}
\usepackage{stfloats}
\usepackage{url}
\usepackage{verbatim}
\usepackage{graphicx}
\def\BibTeX{{\rm B\kern-.05em{\sc i\kern-.025em b}\kern-.08em
    T\kern-.1667em\lower.7ex\hbox{E}\kern-.125emX}}
\usepackage{balance}
\usepackage{amsmath}

\usepackage{booktabs,colortbl}
\usepackage{subfigure}

\usepackage{color}
\usepackage{cite}
\usepackage{makecell}
\usepackage[linesnumbered,ruled]{algorithm2e}
\newtheorem{remark}{Remark}

\usepackage[table,xcdraw]{xcolor} % 必须使用这个包才能为表格添加颜色
\usepackage{tcolorbox}
\usepackage{enumitem}
\usepackage{arydshln}

\begin{document}
\title{LLM-Enhanced Multi-Agent Reinforcement Learning for Unified Electric Vehicles-Charging Station-Grid Optimization in Public Charging Systems}

\author{Yang~Zhang,
        Lindong~Xie,
        Chongyu~Wang,
        Gaojunjie~Li,
        Siqi~Bu,
        and~Edward~Chung~\vspace{-20pt}
\thanks{This work was supported in part by the Innovation and Technology Commission-Mainland-Hong Kong Joint Funding Scheme under Grant MHP/038/23. \textit{(Corresponding Author: Edward Chung.)}}
\thanks{Yang Zhang, Lindong Xie, Gaojunjie Li, Siqi Bu, and Edward Chung are with the Department of Electrical and Electronic Engineering, The Hong Kong Polytechnic University, Hong Kong, China (e-mail: zynolo96@outlook.com; lindong.xie@connect.polyu.hk; gaojunjie.li@polyu.edu.hk; siqi.bu@polyu.edu.hk; edward.cs.chung@polyu.edu.hk).}
\thanks{Chongyu Wang is with the Galvin Center for Electricity Innovation, Illinois Institute of Technology, Chicago, IL 60616, USA (e-mail: chongyu.wang@outlook.com).}
}

\markboth{}%
{How to Use the IEEEtran \LaTeX \ Templates}

\maketitle

\begin{abstract}
In the era of the Internet of Things (IoT), coordinating connected electric vehicle (EV) charging scheduling to balance EV charging satisfaction, station profitability, and smart grid stability presents a complex multi-objective challenge. Existing Multi-Agent Reinforcement Learning (MARL) approaches often struggle with high-dimensional state spaces generated by massive IoT sensing data and conflicting stakeholder interests. This paper proposes a novel LLM-enhanced MARL framework that, for the first time, simultaneously optimizes the Grid, EVs, and Stations within a unified loop. By integrating Large Language Model (LLM), we address two critical bottlenecks: interpretable feature selection and adaptive multi-objective balancing. The LLM analyzes real-time IoT-collected environmental states to extract physically significant features and dynamically assigns weights to conflicting objectives—including profit, user satisfaction, and grid load—using semantic reasoning instead of complex manual tuning. Extensive experiments demonstrate that our framework significantly outperforms state-of-the-art baselines, achieving superior market efficiency while reducing training time by over 70\%. This approach offers a scalable, transparent solution for efficient and sustainable IoT-enabled urban charging infrastructure management.
\end{abstract}

\begin{IEEEkeywords}
Large language model, multi-agent reinforcement learning, feature selection, multi-objective optimization, electric vehicles charging
\end{IEEEkeywords}

\section{Introduction}\label{intro}

\IEEEPARstart{D}{riven} by their environmental benefits and the rapid advancement of Internet of Things (IoT) technologies, connected electric vehicles (EVs) have experienced unprecedented global proliferation over the past decade~\cite{cao2022joint,wang2024demand}. In the United States alone, new EV registrations surged by 40\% to 1.4 million in 2023~\cite{ieaTrendsElectric}, whereas public charging stations increased only by 20\% to 64,187~\cite{consumeraffairsManyCharging}. This widening gap between rapid EV adoption and the relatively slower expansion of public charging infrastructure poses critical challenges in IoT-enabled smart city scenarios, such as meeting escalating charging demands, reducing charging queue concerns, ensuring the economic viability of charging operations, and preventing grid overload. To address these challenges, effective charging scheduling strategies relying on ubiquitous IoT connectivity are essential, encompassing dynamic station pricing, grid load balancing, and real-time management of EV charging power, thus maintaining charging market economics, load security and user satisfaction within the complex IoT ecosystem.

Traditional methods for EV charging scheduling often rely on exact optimization algorithms, formulating the problem through nonlinear programming or mixed-integer linear programming models~\cite{tushar2012economics,yoon2015stackelberg,wang2019stackelberg,zhaoan2020power}. Although these approaches perform well under idealized conditions, they depend critically on accurate mathematical models of the charging environment, a requirement that is challenging to fulfill in dynamic, real-world scenarios~\cite{zhao2024reinforcement}.

Reinforcement learning (RL) has emerged as a promising alternative that eliminates the need for explicit environmental modeling through the trial-and-error interaction paradigm, where agents learn optimal policies directly from experience with the environment. In EV charging contexts, the three key operational aspects, namely station pricing, EV power allocation, and grid power transmission, can each be formulated as a Markov Decision Process (MDP). Various RL algorithms, including deep deterministic policy gradient (DDPG)~\cite{qiu2020deep,li2023constrained}, deep Q-network (DQN)~\cite{liang2020mobility}, proximal policy optimization (PPO)\cite{jiang2021data}, and soft actor-critic (SAC)~\cite{zhang2023safe}, have shown success in optimizing individual aspects. For more complex scenarios involving multiple stakeholders, multi-agent reinforcement learning (MARL) algorithms such as multi-agent DDPG (MADDPG)~\cite{zhang2022multistep,fu2023electric}, multi-agent PPO (MAPPO)~\cite{yang2024multiagent}, and counterfactual multi-agent policy gradient (COMA)~\cite{park2022multi} have demonstrated effectiveness in handling intricate multi-agent interactions.

Despite these advances, a recent survey~\cite{zhao2024reinforcement} reveals a critical gap: existing RL-based studies typically optimize only one or two perspectives within an EV charging market that involves three primary operational dimensions. Specifically, among 111 papers on RL-based EV charging scheduling, none integrate the perspectives of EV users, charging stations, and the grid simultaneously. Such limited perspective can lead to suboptimal or even counterproductive outcomes. For example, optimization strategies focusing solely on EV user satisfaction and charging station revenue may inadvertently exacerbate grid load pressure during peak demand periods, potentially compromising power system security. A comprehensive approach integrating all three perspectives is therefore essential to balance these interconnected factors, enabling adaptive responses to real-time grid conditions, dynamic pricing strategies, and evolving user demands while ensuring system-wide efficiency.

This paper aims to bridge the above gap by integrating all three perspectives, including the EV users, the charging station, and the power grid, into a unified MARL-based framework. To this end, we identify two fundamental challenges. First, the unified framework encompasses an extensive state space; even a scenario featuring a single charging station with 20 charging piles may involve over 100 state features, necessitating efficient feature selection prior to RL deployment. Second, integrating multiple perspectives inevitably yields conflicting stakeholder objectives (e.g., EV users seeking to minimize economic costs versus the charging station aiming to maximize profits), which demands a carefully balanced reward function design. Conventional approaches typically resort to statistical methods for feature selection~\cite{lemhadri2021lassonet,hu2024transforming,ding2005minimum} and multi-objective reinforcement learning (MORL) for balancing conflicting objectives~\cite{lu2023multi, alegre2023sample, yin2024temperature}. Unfortunately, three issues may hinder their applicability in the unified EV charging market: 1) Complexity: Conventional MORL methods often incur high computational costs in seeking optimal trade-offs among conflicting objectives, typically requiring complex mechanisms to approximate or manage a Pareto frontier. Additionally, feature selection based on statistical methods demands extensive parameter tuning and intensive computations. Such complexity severely impedes the rapid deployment and real-time operation required in practical EV charging markets. 
2) Lack of Interpretability: Even when optimal features and balanced weights are derived, these methods often fail to provide clear, intuitive explanations for their determinations, which are crucial for market participants to understand the operational logic behind scheduling decisions.
3) Inadaptability: Most MORL algorithms struggle to dynamically adjust weights in response to real-time market conditions and/or are often limited to balancing only two objectives, failing to accommodate the three distinct stakeholder roles in the unified EV charging market.

Recent advances in large language models (LLMs) present a promising solution to these challenges. Pretrained on vast corpora of general and scientific literature, LLMs have demonstrated superior abilities in understanding, reasoning, and planning~\cite{wu2023brief,chang2024llmscenario,zhao2023survey}. Through comprehending the physical significance of state variables and assessing the urgency and importance of varying stakeholder requirements, LLMs are well-positioned to perform effective feature selection and weight-balancing decisions that align with specific task requirements~\cite{li2025exploring,jeong2024llm}. Compared to traditional feature selection or MORL approaches, LLM-based methods offer several advantages in the unified EV charging market. First, they simplify the process of selecting features and determining weights through prompt-based task descriptions and market state inputs, thus eliminating the need for complex balancing mechanisms or parameter tuning. Second, operating in natural language, LLMs provides transparent, text-based explanations for their feature selection and weight-determination processes. Third, they enable a real-time approach for multi-objective weight assignment, adapting seamlessly to dynamic environmental states~\cite{lu2023multi, alegre2023sample, yin2024temperature}.

Motivated by these advantages, we propose a first LLM-enhanced MARL approach for optimizing unified EV-Station-Grid operations in public charging systems. Our main contributions are as follows:

\textbf{(1) Unified EV-Station-Grid Operation Model:} We propose the first MARL-based unified charging scheduling framework that integrates three perspectives—EVs, the charging station, and the grid—by leveraging queueing theory and Markov game theory. An analysis of the environmental states and reward structures in our model underscores the need for effective feature selection and multi-objective balancing strategies.

\textbf{(2) LLM-based Interpretable Feature Selection:} We leverage LLMs' inherent capability for contextual understanding and semantic reasoning to perform feature selection through intuitive text-based prompts. By translating state features and feature selection task into natural language descriptions, our method allows LLMs to assess feature importance based on their semantic relevance and contribution to stakeholder objectives, providing inherently interpretable justifications. This bypasses the computationally intensive computation or parameter tuning required by statistical methods. An error detection mechanism mitigates potential hallucinations. The top-K features are selected to significantly improve downstream MARL training efficiency.

\textbf{(3) LLM-based Adaptive Multi-objective Balancing:} Similarly, we exploit LLMs' aptitude for reasoning about trade-offs from textual descriptions. Using real-time state descriptions and optimization objective prompts, LLMs dynamically assign interpretable weights to conflicting objectives across three stakeholder perspectives, autonomously adapting to changing market conditions without complex Pareto frontier maintenance required by conventional MORL. An error detection mechanism and a status balance identification mechanism are presented to further ensure validity and optimize efficiency by reducing redundant LLM invocations, respectively. Finally, a balanced overall reward function incorporating these dynamic weights is constructed for MARL policy learning, enabling agents to harmonize multi-stakeholder objectives.

\textbf{(4) Enhanced Performance \& Validation of the Integrated Framework:} We integrate the above LLM modules with MADDPG, enabling the efficient learning of high-quality, balanced scheduling policies for the unified market. Comprehensive experiments demonstrate that our framework significantly outperforms state-of-the-art feature selection-based and MORL-based baselines, achieving superior gains in EV user satisfaction, charging station profits, and grid load pressure reduction, thereby markedly improving overall market operation efficiency while reducing policy training cost.

\section{Related Work and Proposed Extensions}\label{sec:rw}

\subsection{RL-Based EVs Charging Scheduling}
Research in RL-based charging scheduling has primarily evolved along three distinct perspectives. From the EV perspective, researchers have employed RL algorithms to optimize charging satisfaction by minimizing economic costs for charging, range anxiety, and travel time~\cite{li2023constrained,zhang2023cooperative,li2021electric}. For instance, Qian~\textit{et al.} developed a DQN-based approach for efficient charging navigation~\cite{qian2019deep}. From the charging station perspective, studies have focused on maximizing service profits and operational efficiency~\cite{guo2022energy,liang2020mobility,qureshi2023dynamic,shalaby2023model}, with notable work by Qiu~\textit{et al.} implementing DDPG with prioritized experience replay for optimal pricing~\cite{qiu2020deep}. Regarding grid management, research has concentrated on peak-load reduction~\cite{jiang2021data,zhang2023safe,hossain2023efficient,sun2021customized}, exemplified by Sultanuddin~\textit{et al.}'s Double DQN implementation~\cite{sultanuddin2023development}.

Limited research has attempted to employ MARL algorithms to develop coordinated strategies across two types of stakeholders in the EV charging market. Zhang~\textit{et al.} employed MADDPG to jointly optimize grid power transmission and EV charging decisions~\cite{zhang2022multistep}. Fu~\textit{et al.}\cite{fu2023electric} and Yang \textit{et al.}\cite{yang2024multiagent} addressed the game-theoretic interaction between station pricing and EV scheduling, while Park \textit{et al.} implemented COMA and MADDPG to optimize the station energy purchasing and EV charging/discharging~\cite{park2022multi}. However, the integration of all three perspectives remains unexplored, primarily due to challenges in managing high-dimensional state spaces and balancing multiple competing objectives, which are addressed in this paper. For a more detailed review of RL-based EV charging scheduling approaches, readers are referred to a recent survey~\cite{zhao2024reinforcement}.

\subsection{Feature Selection}

Feature selection techniques aim to identify the most relevant features from a candidate set for downstream tasks. Traditional approaches are categorized into three methodologies: filter, wrapper, and embedded methods. Filter methods rely on statistical correlation criteria, including mutual information (MI)\cite{lewis1992feature}, minimum redundancy maximum relevance (mRMR)\cite{ding2005minimum}, Fisher score~\cite{gu2011generalized}, and maximum mean discrepancy~\cite{song2012feature}. Wrapper methods employ heuristic search strategies like sequential selection~\cite{luo2014sequential} and recursive feature elimination (RFE)\cite{guyon2002gene}, to identify optimal feature subsets that maximize model performance. Embedded methods integrate feature selection within the model learning process, utilizing regularization techniques to promote feature sparsity\cite{yuan2006model}. However, these techniques typically incur substantial computational burdens as they require exhaustive statistical computations or large-scale hyperparameter tuning, while also offering limited interpretability for electric vehicle charging scheduling applications.

The advent of LLMs has introduced a novel and simple paradigm for feature selection. By leveraging their semantic understanding capabilities, LLMs can identify task-relevant features through interpretable reasoning about feature significance and task requirements~\cite{choi2022lmpriors}. This approach has shown promising results in healthcare~\cite{li2025exploring,yang2024ice,jeong2024llm} and finance~\cite{han2024large}. Our work extends this paradigm to the interdisciplinary domain of transportation and power systems, and further pioneers a novel framework that integrates LLM-based feature extraction with MARL optimization for intelligent EV charging scheduling.

\subsection{Multi-Objective Reinforcement Learning}

MORL extends standard RL to scenarios involving multiple, often conflicting optimization objectives. MORL algorithms typically aim to find balanced policies that represent different trade-offs among objectives. Early approaches, such as multi-objective Q-learning, addressed this challenge by converting vector-valued Q-values into scalar quantities for action selection~\cite{van2013scalarized}. To identify optimal solutions along the Pareto front (PF) defined as the solution set where improving one objective necessarily compromises another, researchers developed Pareto Q-learning (PQL)\cite{van2014multi} and concave-augmented Pareto Q-learning (CAPQL)\cite{lu2023multi} for tabular settings and deep learning environments, respectively. Further advances include prediction-guided MORL (PGMORL)\cite{xu2020prediction}, which incorporates evolutionary algorithms for efficient Pareto set approximation, and generalized policy improvement-prioritized Dyna (GPI-PD)\cite{alegre2023sample,barreto2017successor}, which constrains solutions to a convex coverage set for more compact representation while maintaining solution quality.

Recent years have witnessed a surge in MORL applications across transportation and power domains, including photovoltaic generation~\cite{yin2024temperature}, charging station pricing~\cite{adetunji2023two}, and battery charging~\cite{xiong2023multiobjective}. While existing MORL approaches have extensively investigated theoretical properties such as algorithmic efficiency and convergence, they often face practical challenges in complex, real-world deployment—specifically: computational complexity from managing Pareto trade-offs, limited interpretability, and inflexibility in dynamic multi-stakeholder balancing. To address the above concerns, we propose an LLM-based multi-objective balance module in this paper. Our LLM-enhanced framework avoids complex mathematical trade-off formulation. It translates objectives and environmental states into semantic representations, enabling dynamic adaptation to real-time multi-objective trade-off requirements. Crucially, it provides natural language justifications, enhancing interpretability.

\section{System Model and Problem Formulation}\label{sec:model}

\begin{figure}[!t]
\centering
\includegraphics[width=3.5in]{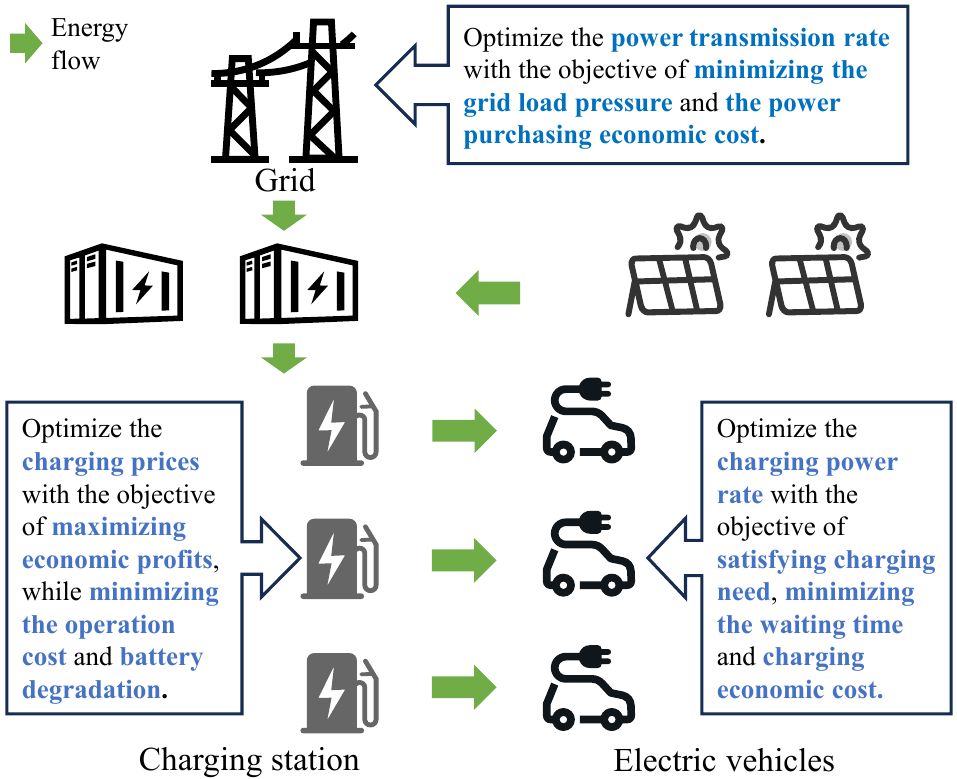}
\caption{Unified EV charging scheduling framework among three perspectives: pricing by the charging station, charging power decisions by EVs, and transmission power control by the grid operator.}
\label{fig:problem}
\vspace{-5mm}
\end{figure}

\subsection{System Model}

As illustrated in Figure~\ref{fig:problem}, we consider a public charging scenario comprising three primary entities: EVs, a charging station equipped with photovoltaic (PV) panels and an energy storage system (ESS), and the power grid. The charging station can supplement its renewable energy generation with energy purchased from the grid to meet charging demands of EVs.

To model the limited charging infrastructure scenario, we adopt an $M/M/s$ queuing system~\cite{bhat2008introduction}, where the first $M$ denotes Markovian arrival process, the second $M$ represents Markovian service times, and $s$ indicates the total number of charging piles. The EV arrival rate $\lambda$ follows a homogeneous Poisson Process~\cite{zhao2023optimal}, while the service rate $\mu$ depends on available charging piles and EV charging power. With occupancy rate $\rho=\lambda/\mu$, the steady-state distribution $\Gamma$ of the queuing process is:

\begin{equation}\label{eqn:que}
\Gamma \left( k \right)=\left\{\begin{array}{ll}
\frac{1}{k!} \rho^{k} {\Gamma \left( 0 \right)}, & \text { for } 0 \leq k < s \\
\frac{\rho^{s}}{s!s^{k-s}} {\Gamma \left( 0 \right)}, & \text { for } k \geq s
\end{array}\right.
\end{equation}
where $k$ denotes the queue length and ${\Gamma \left( 0 \right)}$ is the probability of empty queue defined as:

\begin{equation}\label{eqn:que_ini}
\Gamma(0)=\left(\sum_{k=0}^{s-1} \frac{\rho^{k}}{k!}+\frac{\rho^{s}}{s!} \frac{s}{(s-\rho)}\right)^{-1}
\end{equation}

\begin{remark}
This study focuses exclusively on charging operations to maximize service availability for EV users, excluding vehicle-to-grid (V2G) scenarios that could exacerbate queuing anxiety.
\end{remark}

\subsection{Problem Formulation}\label{problem}

We propose a unified optimization framework that combines the perspectives of EV users, the charging station, and the power grid. The framework operates over a discretized time horizon of $T$ intervals. In this subsection, we first formulate the individual optimization problems for each stakeholder before presenting the unified framework.

\subsubsection{EV User Perspective}

For each time interval $[t,t+1]$, from the perspective of $m$ EV users receiving charging services, the optimization problem can be formulated as follows:

\begin{equation}\label{eqn:ev1}
  \begin{aligned}
  \min_{\mathbf{w}_t=\{w_t^i\}_{i=1}^m} & \Bigg\{ 
  \sum_{i=1}^m \Big(p_t^c w_t^i \Delta t + \alpha_c^i (t - t_a^i)\mathbb{I}\{t=t_{des}^i\} \\
    & + \alpha_s^i (SoC_{des}^i - SoC_t^i)\mathbb{I}\{t=t_{des}^i\}\Big)
  \Bigg\}
  \end{aligned}
\end{equation}
\text{subject to:}
\begin{align}
  & SoC_{t+1}^i = SoC_t^i + \frac{\eta^i w_t^i \Delta t}{E_{ev}^i}, && \forall i \in \{1,\ldots,m\}\label{eqn:ev3} \\
  & SoC_{min}^i \le SoC_t^i \le SoC_{max}^i,&& \forall i \in \{1,\ldots,m\} \label{eqn:ev31} \\
  & w_{min}^i \leq w_t^i \leq w_{max}^i, && \forall i \in \{1,\ldots,m\} \label{eqn:ev4} \\
  & t_a^i \leq t \leq t_{des}^i, && \forall i \in \{1,\ldots,m\} \label{eqn:ev5}
\end{align}

The objective function~(\ref{eqn:ev1}) minimizes the total cost for EV users, comprising three essential components: charging economic costs, waiting time penalties, and state of charge (SoC) dissatisfaction penalties. $p_t^c$, $w_t^i$, $t_a^i$, $t_{des}^i$, $SoC_{des}^i$, and $SoC_t^i$ denote the charging prices at time $t$, the charging power for EV $i$ at time $t$, arrival time for EV $i$, departure time for EV $i$, desired SoC when departure for EV $i$, and the real-time SoC for EV $i$ at time $t$. $\alpha_c^i$ and $\alpha_s^i$ represent the importance coefficients for the two penalty terms in the optimization objective. The indicator function $\mathbb{I}\{t=t_{des}^i\}$ is defined as $1$ if the condition $t=t_{des}^i$ is satisfied, and $0$ else.

Constraint~(\ref{eqn:ev3}) captures the SoC evolution dynamics, incorporating charging efficiency $\eta^i$ and battery capacity $E_{ev}^i$ for EV $i$. 
Constraint~(\ref{eqn:ev31}) enforces operational boundaries $SoC_{min}^i$ and $SoC_{max}^i$ on the battery SoC to protect battery health and longevity. Constraint~(\ref{eqn:ev4}) establishes the physical limitations of charging power, accounting for both minimum $w_{min}^i$ and maximum $w_{max}^i$ charging power. Constraint~(\ref{eqn:ev5}) defines the temporal charging window between the EV's arrival time $t_a^i$ and desired departure time $t_{des}^i$.

\subsubsection{Charging Station Perspective}

During each time interval $[t,t+1]$, the pricing problem from the charging station standpoint is formulated as follows:

\begin{equation}\label{eqn:cs1}
  \max_{p_t^c} \left\{ \sum_{i=1}^m (p_t^c w_t^i \Delta t) - Deg - C_{op} \right\}
\end{equation}
subject to:
\begin{align}
  & Deg = \frac{c_b E_{cs} + c_L}{L_c E_{cs}} E_t^c \label{eqn:cs2} \\
  & E_t^c = \sum_{i=1}^m w_t^i \Delta t \label{eqn:cs3} \\
  & C_{op} = c_{fixed} + c_{var} \cdot E_t^c \label{eqn:cs5} \\
  & p_t^{min} \leq p_t^c \leq p_t^{max} \label{eqn:cs4}
\end{align}

The objective function (\ref{eqn:cs1}) of the charging station comprises three components: maximizing the economic profits through charging service provision ($\sum_{i=1}^m (p_t^c w_t^i \Delta t)$ ), minimizing the battery degradation cost of ESS ($Deg$), and minimizing the charging station operational cost ($C_{op}$).

Constraint~(\ref{eqn:cs2}) models the battery degradation cost as a function of battery cost coefficient $c_b$, labor cost coefficient $c_L$, cycle life $L_c$, and battery size of ESS $E_{cs}$. Constraint~(\ref{eqn:cs3}) calculates the total energy delivered to all EVs during the given time interval as $E_t^c$. Constraint~(\ref{eqn:cs5}) presents the operational costs comprising the fixed components ($c_{fixed}$) and linear components ($c_{var}$) that scale proportionally with energy throughput. Finally, Constraint~(\ref{eqn:cs4}) establishes pricing boundaries $p_t^{min}$ and $p_t^{max}$ that ensure both commercial viability and market competitiveness.

\subsubsection{Grid Perspective}

During each time interval $[t,t+1]$, the power transmission problem from the grid standpoint is formulated as follows:

\begin{equation}\label{eqn:grid1}
  \min_{w_t^g} \left\{M_{load} \left(\frac{w_t^g}{w_{base}^g}\right)^2 - p_t^g w_t^g \Delta t \right\}
\end{equation}
subject to
\begin{align}
  & 0 \leq w_t^g \leq w_t^{max} \label{eqn:grid2} \\
  & E_{t+1}^b = E_t^b + w_t^g \Delta t + w_t^{pv} \Delta t - \sum_{i=1}^m w_t^i \Delta t \label{eqn:grid3} \\
  & -R_D \leq w_t^g - w_{t-1}^g \leq R_U \label{eqn:grid4}\\
  & E_{min}^b \leq E_{t+1}^b \leq E_{max}^b \label{eqn:grid5}
\end{align}

The grid objective in Eq.~(\ref{eqn:grid1}) minimizing the grid load pressure and the total economic cost for the station purchasing power from the grid. The first term introduces a quadratic function that discourages excessive load variations, thereby promoting smoother grid operation~\cite{tuchnitz2021development}, while the second term quantifies the direct electricity purchasing cost at grid price $p_t^g$. Parameters $M_{load}$ and $w_{base}^g$ denote the load management factor and the base power from the grid, respectively. While the power purchasing cost can be conceptually included in the charging station's optimization sub-problem, we incorporate it into the grid's objective function since the decision variable, transmission power $w_t^g$, appears in the grid-side constraints and directly affects the load burden of the grid.

Constraint~(\ref{eqn:grid2}) establishes limits $w_t^{max}$ on instantaneous power draw from the grid, respecting infrastructure capabilities. Constraint~(\ref{eqn:grid3}) maintains equilibrium between power supply (transmission power $w_t^g$ and PV generation $w_t^{pv}$) and EV charging demand for the stored energy $E_t^b$ in the battery system of the charging station. Constraint~(\ref{eqn:grid4}) implements essential ramping constraints with downward $R_D$ and upward $R_U$ limits to prevent rapid load fluctuations that could destabilize the grid. Constraint~(\ref{eqn:grid5}) enforces appropriate energy storage boundaries $E_{min}^b$ and $E_{max}^b$ for the charging station.

\subsubsection{Unified Optimization Framework}\label{sec:uni}

By integrating the above three perspectives, we formulate a unified optimization problem that balances the interests of all stakeholders:

\begin{equation}\label{eqn:int1}
\begin{aligned}
  \min_{\{w_t^g,\mathbf{w}_t,p_t^c\}} & c_{ev} \Bigg\{ 
  \sum_{i=1}^m \Big(p_t^c w_t^i \Delta t + \alpha_c^i (t - t_a^i)\mathbb{I}\{t=t_{des}^i\} \\
    & + \alpha_s^i (SoC_{des}^i - SoC_t^i)\mathbb{I}\{t=t_{des}^i\}\Big)
  \Bigg\}\\
    & - c_{cs}\left\{ \sum_{i=1}^m (p_t^c w_t^i \Delta t) - Deg - C_{op}\right\}\\
    & + c_{g}\left\{ p_t^g w_t^g \Delta t + M_{load} \left(\frac{w_t^g}{w_{base}^g}\right)^2 \right\}
\end{aligned}
\end{equation}
\begin{flalign}
& \text{subject to} \ (\ref{eqn:ev3}) - (\ref{eqn:ev5}), (\ref{eqn:cs2}) - (\ref{eqn:cs4}), (\ref{eqn:grid2}) - (\ref{eqn:grid5}) & 
\end{flalign} 
\noindent where $c_{ev}$, $c_{cs}$, and $c_{g}$ denote weights assigned to the optimization objectives of EV users, the charging station, and the grid in the unified optimization problem, respectively. 

\section{Markov Game model}\label{sec:mg}

Although the theoretical model proposed in Section~\ref{problem} demonstrates the potential for deriving optimal charging schedules, its practical implementation faces two major challenges: the difficulty in obtaining accurate users' objective formulation due to privacy restrictions, and the inherent uncertainties of model parameters in dynamic real-world environments. To mitigate this issue, we propose an LLM-enhanced MARL optimization approach that develops optimal charging strategies in a data-driven manner. Before introducing our approach, this section formulates the EV charging scheduling as a Markov Game model and analyzes its key components.

\subsection{Markov Game Model Definition}\label{sec:markov}

The EV charging market is modeled as a Markov Game \(\Gamma = \langle \mathcal{I}, \mathcal{S}, \{\mathcal{A}^i\}_{i \in \mathcal{I}}, \mathcal{T}, \mathcal{R}, \gamma \rangle\), where:

\textbf{Agents (\(\mathcal{I}\)):}  
The agent set \(\mathcal{I} = \{\mathbf{V}, CS, G\}\) comprises three types of decision-making entities: the group of charged EVs \(\mathbf{V} = \{V_1, V_2, \ldots, V_m\}\), the charging station (\(CS\)), and the grid operator (\(G\)).

\textbf{State Space (\(\mathcal{S}\)):}  
The state \(s \in \mathcal{S}\) encapsulates comprehensive features about the charging market, including:
\begin{itemize}
  \item \textbf{EV charging requirements:} 
  \begin{itemize}
  \item Desired SoC \(\mathbf{SoC}_{des} = \{SoC_{des}^i\}_{i=1}^m\), 
  \item Desired departure times \(\mathbf{t}_{des} = \{t_{des}^i\}_{i=1}^m\), 
  \item EV-specific cost coefficients \(\boldsymbol{\alpha}_s = \{\alpha_{s}^i\}_{i=1}^m\) (unsatisfied SoC cost) and \(\boldsymbol{\alpha}_c = \{\alpha_{c}^i\}_{i=1}^m\) (charging time cost)
  \item EV charging efficiencies \(\boldsymbol{\eta} = \{\eta^i\}_{i=1}^m\)
  \end{itemize}
  \item \textbf{Station metrics:}
    \begin{itemize}
      \item Current time \(t\)
      \item Number of charged EVs \(m\)
      \item Energy stored in the station’s battery system \(E_t^b\)
      \item Charging price \(p_{t-1}^c\) last time step and predicted price \(\hat{p}_t^c\) from pricing information last day
      \item PV generation \(w_{t-1}^{pv}\) last time step and predicted PV output \(\hat{w}_t^{pv}\) from the data last day 
      \item Total charging demand \(E_{t-1}^c\) last time step and predicted charging demand \(\hat{E}_t^c\) from the data last day
      \item Current queue length \(k_t\) and predicted queue length \(\hat{k}_t\) based on Eq.~(\ref{eqn:que})
      \item Charging station parameters (\(c_b\), \(c_L\), \(L_c\), \(E_{cs}\))
    \end{itemize}
  \item \textbf{Grid metrics:} 
      \begin{itemize}
      \item Base grid load \(w_{base}^g\) 
      \item Grid price \(p_{t-1}^g\) last time step and predicted grid price \(\hat{p}_t^g\) from the data last day 
    \end{itemize}
\end{itemize}

\textbf{Action Space (\(\{\mathcal{A}^i\}_{i \in \mathcal{I}}\)):}  
At each time step \(t\), the actions available to each agent type are:
\begin{itemize}
  \item \textbf{EVs:} Charging power \(w^i \in \{w_{min}^i, \ldots, w_{max}^i\}\) for each EV \(i\)
  \item \textbf{Charging Station:} Charging price \(p_t^c \in [p_t^{min}, p_t^{max}]\)
  \item \textbf{Grid:} Transmission power \(w_t^g \in [0, w_t^{max}]\)
\end{itemize}

\textbf{Transition Function (\(\mathcal{T}\)):}  
The transition function defines the state evolution based on the current state and taken actions, reflecting the system dynamics.

\textbf{Reward Function ($\mathcal{R}$):} The reward at time $t$ is defined as:
\begin{equation}\label{eqn:13}
r_{t} = c_{t}^{ev}f(r_{t}^{ev}) + c_{t}^{cs}g(r_{t}^{cs}) + c_{t}^{g}h(r_{t}^{g})
\end{equation}
where $f(\cdot)$, $g(\cdot)$, and $h(\cdot)$ are linear normalization functions to shape rewards $r_t^{ev}$, $r_t^{cs}$, $r_t^{g}$ into the interval $[-10,20]$, respectively.
According to the optimization objectives defined in Section~\ref{problem}, the rewards of EVs, the charging station, and the grid are defined as follows: 
\begin{equation}\label{eqn:rew_ev}
\begin{aligned}
r_t^{ev} =  & -\sum_{i=1}^m p_t^c w_t^i \Delta t - \alpha_c \sum_{i=1}^m (t - t_a^i)\mathbb{I}\{t=t_{des}^i\}  \\
& - \alpha_s \sum_{i=1}^m (SoC_{des}^i - SoC_{t_{des}^i}^i)\mathbb{I}\{t=t_{des}^i\} 
\end{aligned}
\end{equation}
\begin{equation}\label{eqn:rew_cs}
r_t^{cs} = \sum_{i=1}^m (p_t^c w_t^i \Delta t) - Deg - C_{op}
\end{equation}
\begin{equation}\label{eqn:rew_csg}
r_t^{g} = -p_t^g w_t^g \Delta t - M_{load} \left(\frac{w_t^g}{w_{base}^g}\right)^2 
\end{equation}

\textbf{Discount Factor ($\gamma$):} $\gamma \in [0,1]$ balances immediate and future rewards in the cumulative objective $R_n = \sum\nolimits_{t = 1}^T \gamma^{t-1}{{r_t}}$.

\begin{figure*}[!t]
\centering
\includegraphics[width=6.5in]{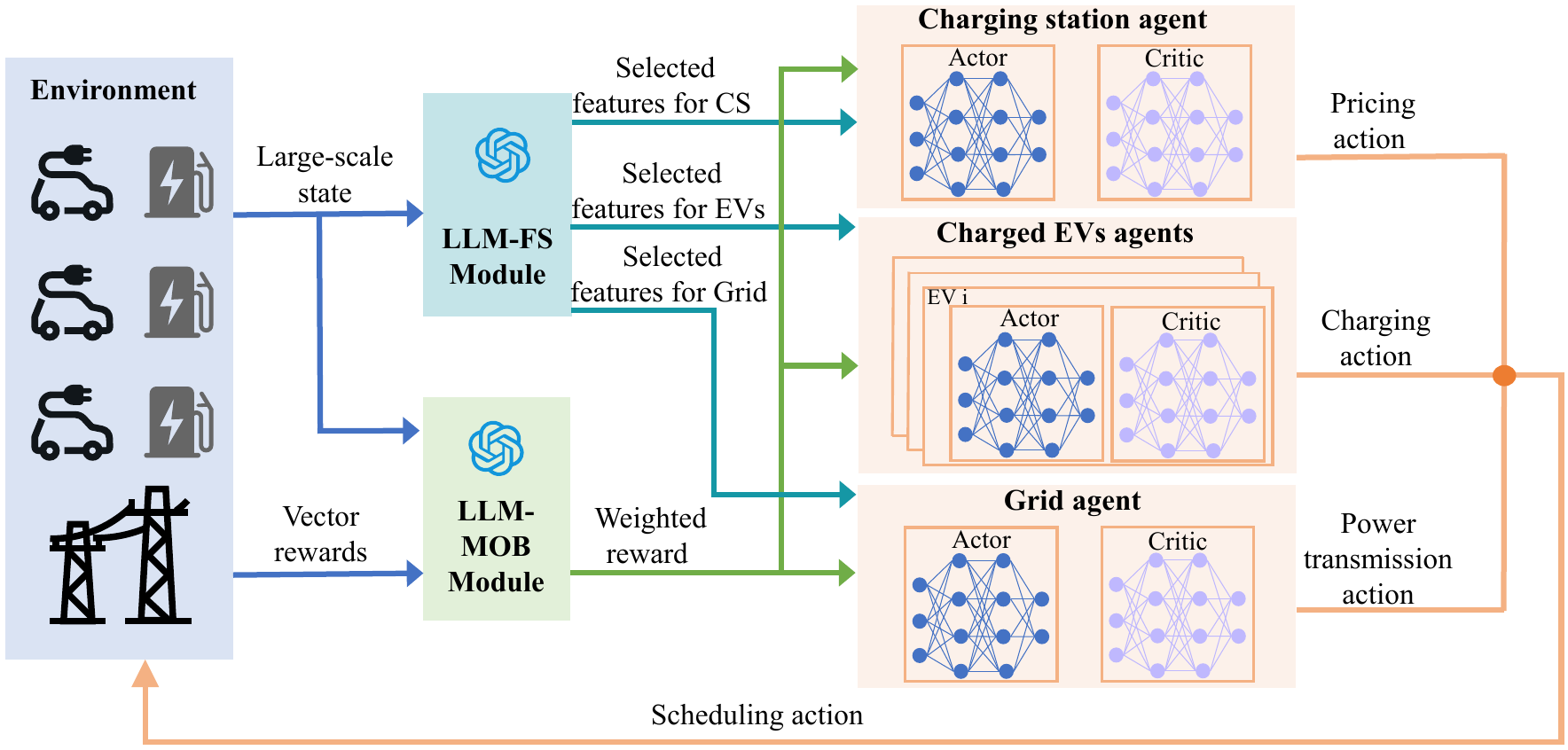}
\caption{Overview of the proposed approach. The framework consists of two LLM-based modules: features selection (LLM-FS) and multi-objective balancing (LLM-MOB). Building upon these modules, we employ MADDPG, a widely-adopted MARL algorithm, to develop the optimal scheduling policy.}
\label{model}
\vspace{-3mm}
\end{figure*}

\subsection{State and Reward Analysis}

The Markov Game formulation highlights two primary challenges for efficient policy development using MARL: the high dimensionality of the state space and the complex interdependence of rewards from multiple stakeholders.

For example, in a charging station with 20 charging piles, the state space reaches a dimensionality of 118, incorporating diverse metrics from EV operations, station management, and grid conditions. Such a high state dimensionality increases the computational complexity and slows down the convergence of MARL algorithms, complicating the derivation of optimal scheduling strategies.

Moreover, the multi-stakeholder nature of the charging scheduling problem requires a careful balancing of reward components. The effectiveness of the overall reward function depends critically on how well the weights adapt to a dynamic charging market, which is characterized by fluctuating charging demands, grid conditions, station utilization, and renewable energy availability.

Addressing these challenges, our approach introduces two key innovations: an interpretable feature extraction method to reduce state dimensionality while preserving critical decision-making information, and an adaptive and interpretable reward weighting method that dynamically adjusts to time-varying environmental conditions. The subsequent sections will detail these methodological contributions, paving the way for a more efficient MARL implementation enhanced by LLM for EV charging scheduling.

\section{Our Approach}\label{sec:app}

In this section, we propose an LLM-enhanced MARL approach to achieve optimal charging scheduling in a unified EV charging scenario. As illustrated in Figure~\ref{model}, our framework comprises three key components: (1) the LLM-based feature selection (LLM-FS) module, which identifies and ranks the most relevant state features while providing interpretable explanations; (2) the LLM-based multi-objective balancing (LLM-MOB) module, which adaptively adjusts the trade-offs among multiple objectives based on the real-time environment; and (3) a MARL module leveraging the above LLM modules, learns the optimal scheduling policy using MADDPG. The following subsections detail each component sequentially.

\subsection{LLM for Feature Selection}

\begin{figure}[!t]
\centering
\includegraphics[width=3.5in]{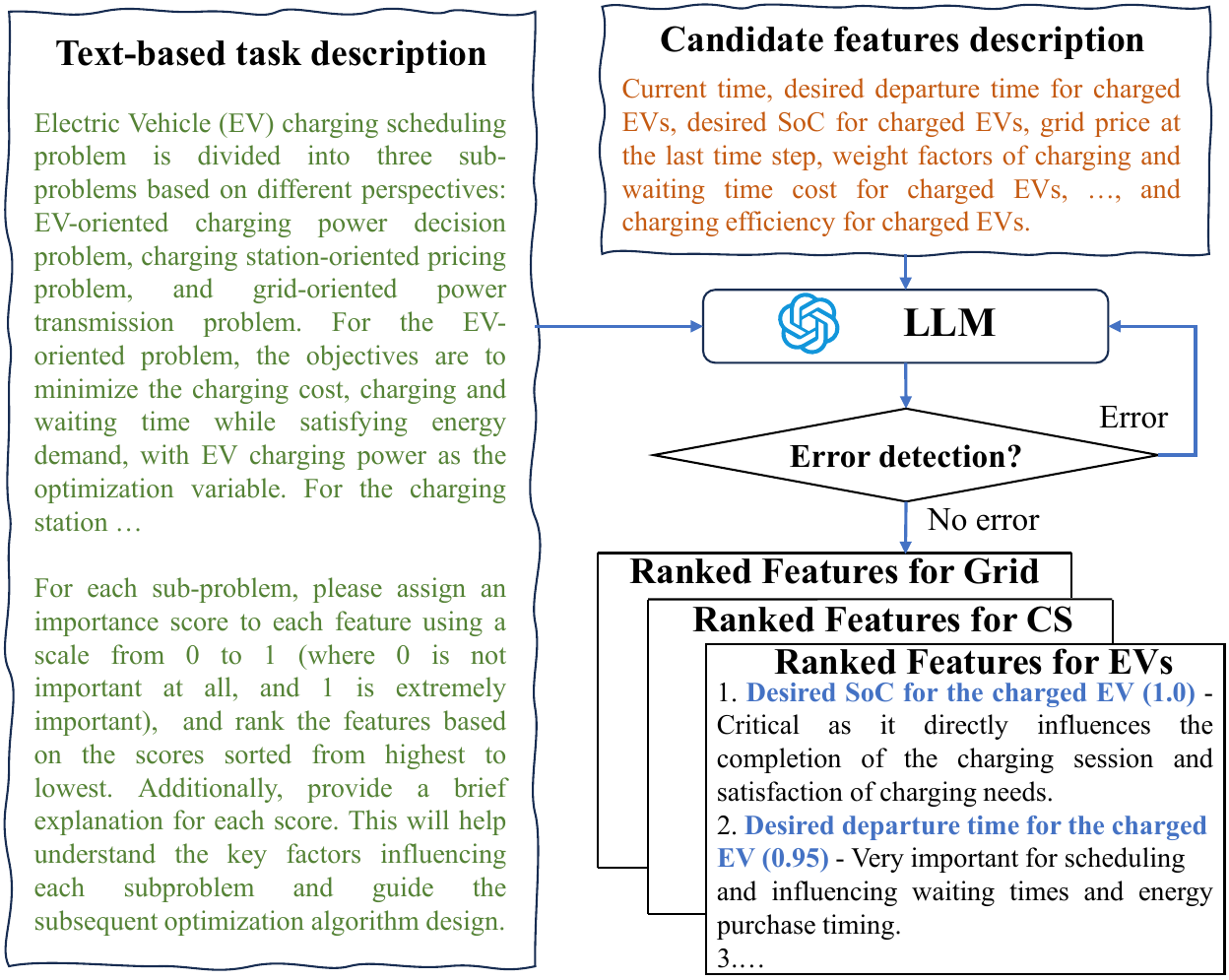}
\caption{LLM-based interpretable feature selection approach.}
\label{fig:states}
\vspace{-3mm}
\end{figure}

In this subsection, we propose an LLM-based interpretable feature selection approach that systematically identifies and ranks the most relevant state features for each optimization subproblem. As illustrated in Figure~\ref{fig:states}, our approach comprises five key steps:

\textbf{Step 1: Feature Selection Task Description Formation.}  
We begin by formulating the feature selection task as a natural language description, denoted as \(Des_{selection}\). This description specifies both the optimization objectives and the corresponding decision variables for each subproblem, considering the perspectives of EVs, the charging station, and the grid. The text-based task description provides the LLM with the necessary context for evaluating feature relevance.

\textbf{Step 2: Candidate Feature Description.}  
We describe all candidate features in natural language as \(Des_{features}\). This description encompasses the entire environmental state of the charging market as detailed in Section~\ref{sec:markov}, enabling the LLM to capture the semantic meaning and potential importance of each feature.

\textbf{Step 3: LLM-based Feature Scoring.}  
Inspired by~\cite{li2025exploring}, We construct a structured prompt by combining \(Des_{selection}\) and \(Des_{features}\), and then feed the prompt to the LLM. For each subproblem, the LLM evaluates feature relevance on a scale of \([0,1]\), where \(1\) indicates maximum importance and \(0\) indicates minimum importance. Crucially, the LLM also provides explicit reasoning for each score, explaining how each feature contributes to the scheduling objectives. For instance, in the EV-oriented charging power decision subproblem depicted in Figure~\ref{fig:states}, the feature “Desired SoC for the charged EV” may receive a high score \(1.0\) because of its direct impact on energy demand. Formally, the scoring process is defined as:
\begin{equation}
    Scores^{f,o}, Reason^{f,o} = \mathcal{M}(Des_{selection}, Des_{features}),
\end{equation}
where \(\mathcal{M}\) denotes the LLM, and \(Scores^{f,o}\) along with \(Reason^{f,o}\) represent the scores and corresponding explanations for subproblem \(o \in \{cs, ev, g\}\). Here, \(cs\), \(ev\), and \(g\) represent the charging station perspective, EV user perspective, and grid perspective, respectively.

\textbf{Step 4: Error Detection.} 
To mitigate potential hallucination errors in the LLM's output during the feature selection task, we implement an error detection mechanism. Specifically, if any importance score in \(Scores^{f,o}\) falls outside the valid range \([0, 1]\), the process reverts to Step 3 with an additional prompt notifying the LLM of the error. This re-evaluation is repeated until all scores fall within the accepted range.

\textbf{Step 5: Feature Selection.}  
We rank the features based on their scores in descending order and select the top \(K\) features for each subproblem, ensuring a concise yet comprehensive state representation. This process is formally expressed as:
\begin{equation}\label{eq:feature_selection}
    \mathbf{F^o} = \text{Top-K}(\mathbf{F}, Scores^{f,o}),
\end{equation}
where $\mathbf{F}$ and \(\mathbf{F}^o\) denote the candidate features listed in Section~\ref{sec:markov} and the selected \(K\) features for each subproblem \(o\), respectively.

The LLM-based feature selection approach offers two key advantages over traditional methods. First, it leverages the domain knowledge encoded within LLMs to discern the physical significance of features, thereby enabling an interpretable and semantically effective selection process. Second, unlike conventional approaches that demand complex statistical computations and extensive parameter tuning, our method provides a flexible and efficient feature selection approach through a streamlined prompting mechanism.

\subsection{LLM for Multi-Objective Balancing}

\begin{figure}[!t]
\centering
\includegraphics[width=3.5in]{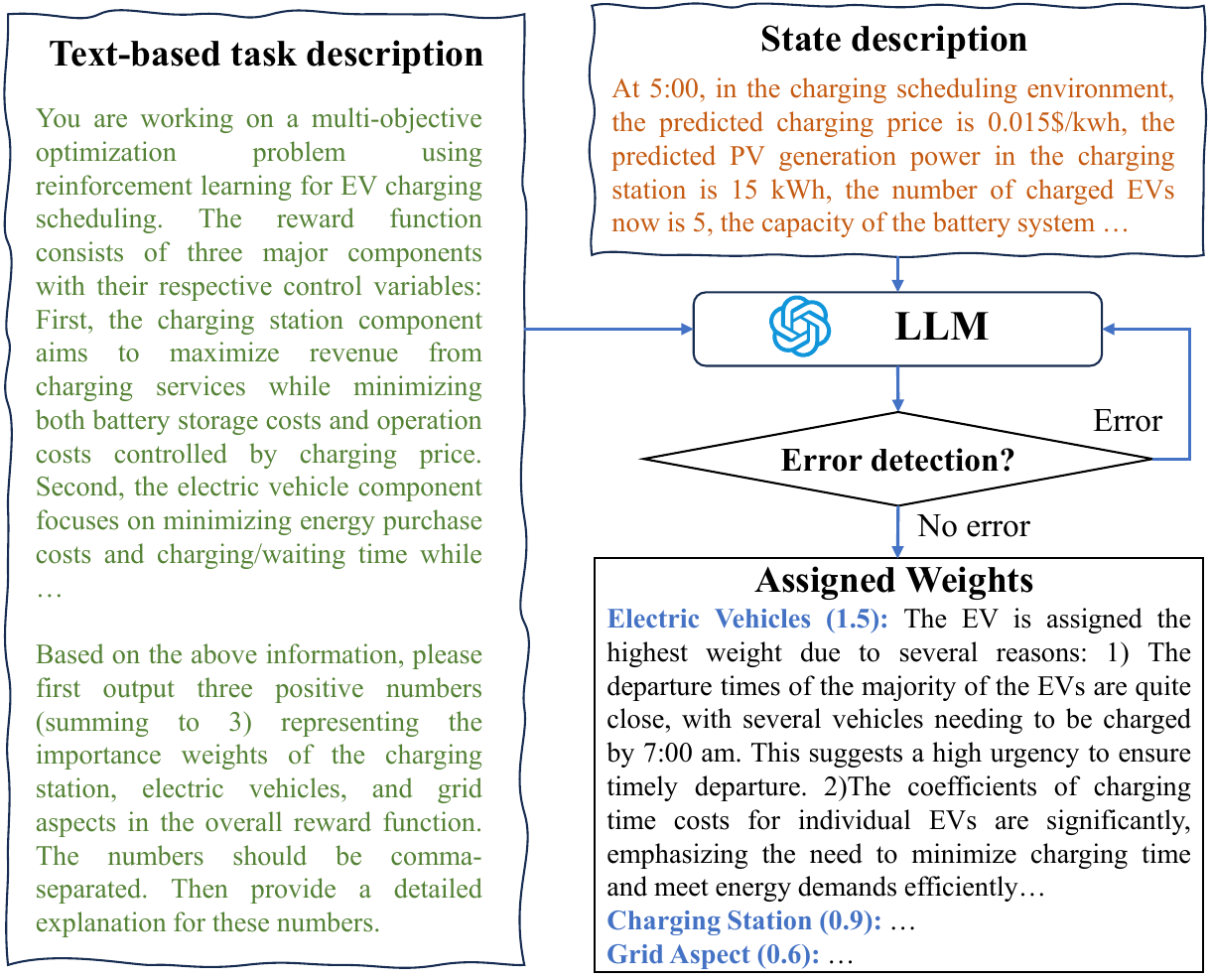}
\caption{LLM-based adaptive and interpretable multi-objective balancing approach.}
\label{fig:weights}
\vspace{-5mm}
\end{figure}

To achieve a dynamic, context-aware balance among multiple objectives in EV charging scheduling, we propose an LLM-based multi-objective balancing module. This module adaptively assigns importance weights to each objective based on the current environmental state. As shown in Figure~\ref{fig:weights}, the process consists of five key steps:

\textbf{Step 1: Multi-Objective Balancing Task Description Formation.}  
We formulate the multi-objective balancing task as a natural language description, \(Des_{balance}\), which details the three key objectives corresponding to the EVs, the charging station, and the grid. The description outlines the potential conflicts and trade-offs among these objectives, providing comprehensive context for weight determination.

\textbf{Step 2: Environmental State Description.}  
We translate the current environmental state into a natural language description, \(Des_{state}\), which includes the real-time information of important features selected by the LLM-FS module. This enables the LLM to understand the current market conditions and make informed decisions about objective prioritization.

\textbf{Step 3: Weight Assignment.}  
We combine \(Des_{balance}\) and \(Des_{state}\) into a structured prompt and feed it to the LLM. The LLM then analyzes the current situation and assigns importance weights \(c^o_t\) for each objective on a scale of \([0,3]\), where higher values indicate greater importance, while ensuring that the weights sum to 3. The LLM also provides explicit reasoning for each weight assignment. Formally, at each time step \(t\), for each perspective \(o \in \{cs, ev, g\}\), the process is represented as:
\begin{equation}\label{eq:weights_balance}
    c^o_t, Reason^{c,o}_t = \mathcal{M}(Des_{balance}, Des_{state}).
\end{equation}
where $Reason^{c,o}_t$ is the reasoning for weight assignment. 

\textbf{Step 4: Error Detection.} 
To prevent potential hallucination errors in the LLM's outputs, we incorporate an error detection mechanism analogous to that used in the LLM-FS module. Specifically, if any weight \(c_t^{o}\) falls outside the valid range \([0, 3]\) or if the sum of weights does not equal 3, the process returns to Step 3 with an additional prompt that notifies the LLM of the detected error. %This re-evaluation is repeated until all weights meet the required criteria.

\textbf{Step 5: Weighted Reward Calculation.}  
Finally, we calculate the unified reward signal for the RL agent using the weights assigned by the LLM, following the weighted sum formulation in Eq.~(\ref{eqn:13}).

Since re-determining weights using the LLM at every training episode can be computationally expensive, we introduce an imbalance identification mechanism. The LLM-MOB module is triggered only when the system status is identified as \textbf{Imbalanced} every \(G\) training episodes. Specifically, the status is classified as \textbf{Imbalanced} if the cumulative reward from one stakeholder exceeds the combined rewards of the other two, i.e., $\sum\nolimits_G\sum\nolimits_{t = 1}^T {r_t^{ev}}  > \sum\nolimits_G\sum\nolimits_{t = 1}^T {(r_t^{cs} + r_t^{g})}$, $\sum\nolimits_G\sum\nolimits_{t = 1}^T {r_t^{cs}}  > \sum\nolimits_G\sum\nolimits_{t = 1}^T {(r_t^{ev} + r_t^{g})}$, or $\sum\nolimits_G\sum\nolimits_{t = 1}^T {r_t^{g}}  > \sum\nolimits_G\sum\nolimits_{t = 1}^T {(r_t^{cs} + r_t^{ev})}$.
If the condition is not met, the previously determined weights are retained.

This LLM-based multi-objective balancing approach offers several benefits: (1) adaptive objective prioritization based on real-time environmental states; (2) interpretable weight assignments through explicit reasoning; and (3) implementation-friendly that eliminates the requirement for complex multi-objective balance mechanisms.

\subsection{LLM-Enhanced MARL Method}

Building upon the LLM-FS and LLM-MOB modules, we develop a comprehensive LLM-enhanced MARL framework that proceeds in two phases: a training phase for learning the optimal charging scheduling policy, and an execution phase for deploying the learned policy in real-time operations.

\subsubsection{Training Phase}

We adopt MADDPG as our core MARL algorithm for three reasons: First, it supports continuous actions (pricing and power transmission decisions) as well as discrete actions (charging power decisions), Second, it enables coordinated policy learning among heterogeneous agents. Third, it ensures training stability via centralized training with deployment flexibility by decentralized execution.

Under MADDPG, each agent \(i \in \mathcal{I}\) maintains an actor network \(\mu^i\) with parameters \(\theta^i\) and a critic network \(Q^i\) with parameters \(\phi^i\). The actor maps the selected feature values \(f^i_t \in \mathbf{F}^o\) to actions \(a^i_t\), as given by:
\begin{equation}\label{eq:action}
    a^i_t = \mu^i(f^i_t) + \mathcal{N}_e(0,\zeta^e),
\end{equation}
where \(\mathcal{N}_e\) denotes Gaussian exploration noise with episode-dependent variance \(\zeta^e\). $e$ denotes the training episode number.

The critic evaluates the state-action value \(Q^i(s_t, a_t)\) using the global state \(s_t\) and the joint action \(a_t\), which concatenates selected features and actions of all agents, respectively.

\begin{algorithm}[!t]
{\small
\caption{{LLM-Enhanced MARL for unified EVs Charging Scheduling}}\label{alg1}
\KwIn{Training episode count \(E_{train}\), execution episode count \(E_{exe}\), time steps per episode \(T\)}
\KwOut{Optimal charging scheduling policy}
%Initialize replay buffer \(\mathcal{D}\)\\
\textcolor{gray}{// Feature Selection Process}\\
Construct a prompt that includes the feature selection task description \(Des_{selection}\) and candidate feature descriptions \(Des_{features}\).\\
Feed the prompt into the LLM to obtain feature scores and explanations using Eq.~(\ref{eq:feature_selection}).\\
Select the Top-\(K\) candidate features based on the scores.\\
\textcolor{gray}{// Training Phase}\\
\For{each training episode \(e = 1\) \KwTo \(E_{train}\)}{
    \For{each time step \(t = 1\) \KwTo \(T\)}{
        Observe the current state of the environment.\\
        Sample an action according to Eq.~(\ref{eq:action}).\\
        Execute the action and observe the vector reward.\\
        \textcolor{gray}{// Multi-Objective Reward Balancing}\\
        \eIf{system status is \textbf{Imbalanced} and \(e \bmod G = 0\)}{
            Construct a prompt with the multi-objective task description \(Des_{balance}\) and current state description \(Des_{state}\).\\
            Feed the prompt into the LLM to obtain weights and corresponding explanations using Eq.~(\ref{eq:weights_balance}).\\
            Calculate the weighted rewards using these generated weights.\\
        }{
            Calculate the weighted rewards using the previously determined weights.\\
        }
        Update the actor network using Eq.~(\ref{eq:train_actor}).\\
        Update the critic network using Eq.~(\ref{eq:train_critic}).\\
    }
}
\textcolor{gray}{// Execution Phase}\\
\For{each execution episode \(e = 1\) \KwTo \(E_{exe}\)}{
    \For{each time step \(t = 1\) \KwTo \(T\)}{
        Observe the current state of the environment.\\
        Generate an action using the trained actor network.\\
        Execute the action in the environment.\\
    }
}
}
\end{algorithm}

During training, the actor of agent \(i\) is updated according to the gradient:
\begin{equation}\label{eq:train_actor}
\nabla_{\theta^i} J\big(\mu^i\big) = \mathbb{E}_{s_t,a_t \sim \mathcal{D}} \Big[\nabla_{\theta^i}\mu^i(f_t^i)\nabla_{a_t^i}Q^i(s_t,a_t)\big|_{a_t^i=\mu^i(f_t^i)}\Big],
\end{equation}
where \(\mathcal{D}\) denotes the replay buffer.

The critic for agent \(i\) is updated by minimizing the loss:
\begin{equation}\label{eq:train_critic}
    \begin{aligned}
    &\mathcal{L}(\phi^i) = \mathbb{E}_{s_t,a_t \sim \mathcal{D}}\Big[(Q^i(s_t,a_t) - y_t)^2\Big]\\
    &y_t = r_t + \gamma {Q^i}'\Big(s_{t+1},a'_{t+1}\Big)\Big|_{a^{i'}_{t+1}={\mu^i}'(f_{t+1})},
    \end{aligned}
\end{equation}
where $r_t$, $\mu'$, and $Q'$ represent the weighted reward, target actor network and target critic network, respectively. The target networks are slowly updated to track their learned actor or critic networks using parameter \(\epsilon \in [0,1]\).

\subsubsection{Execution Phase}

Following centralized training, the execution phase deploys each agent in a decentralized manner where actor networks service different shareholders independently. Each agent \(i\) generates the action based solely on its local observations, i.e., the selected feature values, utilizing the trained actor network without exploration noise: $a^i_t = \mu^i(f^i_t)$. This decentralized execution framework preserves the benefits of centralized training while ensuring scalability and responsive real-time decision-making.

Algorithm~\ref{alg1} summarizes the complete operation process. The algorithm starts with the LLM-FS module selecting relevant features (lines 2–4). During training episodes, the agents interact with the environment (lines 8–10), while the LLM-MOB module adaptively adjusts the objective weights when the system is identified as \textbf{Imbalanced} (lines 12–15). The actor and critic networks are then updated based on the weighted rewards (lines 19–20). In the execution phase, agents use their trained actor networks to generate decisions in a decentralized manner (lines 24–30).

\section{Experiments Analysis}\label{sec:exp}

\subsection{Experiments Methodology}

\subsubsection{Environmental Settings}

\begin{table}[!t]
\centering
\caption{Parameter Settings for EV Charging Market}
\label{tab:paras_ev}
\begin{tabular}{lc}
\toprule
\textbf{Parameter} & \textbf{Value} \\
\midrule
Minimum charging price ($p_{t}^{min}$) & $[7.5, 15]$ \$/MWh (seasonal) \\
Maximum charging price ($p_{t}^{max}$) & $[35, 75]$ \$/MWh (seasonal) \\
Grid price ($p_{t}^g$) & $[5.5, 17.5]$ \$/MWh (seasonal) \\
Minimum charging power ($w_{min}^i$) & $[7.2, 12.5]$ kW (EV-dependent) \\
Maximum charging power ($w_{max}^i$) & $[29.0, 50.0]$ kW (EV-dependent) \\
Target SoC ($SoC_{des}^i$) & $[70\%, 95\%]$ (EV-dependent) \\
Arrival time ($t_{c}^i$) & $[0, 23]$ h (EV-dependent) \\
Initial SoC ($SoC_{t_c^i}^i$) & $[10\%, 50\%]$ (EV-dependent) \\
Battery capacity ($E_{ev}^i$) & $[72.6, 125]$ kWh (EV-dependent) \\
Latest departure ($t_{des}^i$) & $[1, 24]$ h (EV-dependent) \\
Maximum transmission power ($w_t^{max}$) & 2000 kWh \\
Transmission limit ($R_U$ and $R_D$) & 500 kWh\\
Load management factor ($M_{load}$) & 10 \\
Charging efficiency ($\eta$) & 0.95 \\
Battery cost ($c_b$) & 40 \$/kWh \\
Battery replacement labor ($c_L$) & 20 \$ \\
Battery lifecycle ($L_c$) & 25000 cycles \\
Weight of charging time cost ($\alpha_c$) & $[0, 1]$ (EV-dependent) \\
Weight of dissatisfied energy ($\alpha_s$) & $[0, 1]$ (EV-dependent) \\
Base load from grid ($w_{base}^g$) & 1000 \\
EV arrival rate ($\lambda$) & $[5, 30]$ (time-dependent) \\
CS service rate ($\mu$) & $[2.5,15]$ (time-dependent) \\
Number of charging piles ($s$) & 20 \\
Number of selected features ($K$) & 5 \\
\bottomrule
\end{tabular}
\vspace{-3mm}
\end{table}

Our experimental evaluation utilizes comprehensive real-world data from Pennsylvania—New Jersey—Maryland (PJM)~\cite{pjmWebsite}, a major electricity transmission organization serving 65 million customers across 13 U.S. states. We employ hourly electricity prices and solar generation data from April 2023 to March 2024, where the first 20 days of each month are dedicated to model training, while the remaining days serve for execution and performance evaluation. The electric vehicle specifications in our study are based on the top 10 best-selling EVs in the 2023 U.S. market~\cite{Cleantechnica}. Vehicle arrivals follow a Poisson process distribution~\cite{zhao2023optimal}, with our simulation environment handling an average daily throughput exceeding 200 EVs. Detailed parameter specifications are outlined in Table~\ref{tab:paras_ev}.

\subsubsection{Algorithm Benchmark} We compare our approach with two groups of methods that emphasize feature selection and multi-objective balancing, respectively.

\textbf{Group 1: Feature Selection-based Methods.}  
We evaluate five feature selection methods: LassoNet~\cite{lemhadri2021lassonet}, RFE~\cite{guyon2002gene}, MI~\cite{lewis1992feature}, mRMR~\cite{ding2005minimum}, and Transformer~\cite{hu2024transforming}. Each of these methods is incorporated into two MARL configurations: (1) MADDPG with fixed weights, denoted as “\texttt{x+MARL}”, where “\texttt{x}” represents the feature selection method. (2) MADDPG empowered by LLM-MOB module, denoted as “\texttt{x+LLM-MOB}”.

\textbf{Group 2: MORL-based Methods.}  
We assess three MORL approaches: CAPQL~\cite{lu2023multi}, GPI-PD~\cite{alegre2023sample}, and PGMORL~\cite{xu2020prediction}. For each method, we test two feature selection configurations: (1) Using the complete feature set, denoted as “\texttt{x}”, where “\texttt{x}” corresponds to the MORL method. (2) Using the feature subset selected by the LLM-FS module, denoted as “\texttt{x+LLM-FS}”.

In addition, a baseline "\texttt{Vanilla MARL}", i.e, MADDPG with the complete feature set and fixed weights is included.

\subsubsection{Algorithm Settings}

Our implementation utilizes GPT-4o~\cite{hurst2024gpt} (released by OpenAI in May 2024) as the LLM for both feature selection and multi-objective balancing. Ablation studies further investigate the performance with alternative LLMs. Each training and execution episode consists of 24 time steps (representing one day). The RL algorithm parameters are standardized across all benchmarks, as detailed in Table~\ref{tab:paras_rl}.

\begin{table}[!t]
\centering
\caption{Parameter Settings for Examined Algorithms}
\label{tab:paras_rl}
\begin{tabular}{lc}
\toprule
\textbf{Parameter} & \textbf{Value} \\
\midrule
Training episodes & 10,000 \\
Batch size & 128 \\
Replay buffer size & 10,000 \\
Learning rate & 0.0001 \\
Actor network layers & 3 \\
Critic network layers & 2 \\
Neurons per layer & 256 \\
Discount factor ($\gamma$) & 0.95 \\
Target network update rate ($\epsilon$) & 0.005 \\
Initial Gaussian noise & 1 \\
Minimum Gaussian noise & 0.01 \\
Episodes interval for status balance identification ($G$) & 100 \\
\bottomrule
\end{tabular}
\vspace{-3mm}
\end{table}

\subsubsection{Evaluation Metrics}

In addition to the \textbf{episodic weighted rewards $R_n$} defined in Section~\ref{sec:markov} for evaluating training performance, we adopt five metrics which capture diverse stakeholder perspectives during the execution phase:
\begin{itemize}
    \item \textbf{daily profits of the charging station (DPS):} the economic returns of the charging station defined as the charging revenue minus the sum of electricity purchase and operational costs over 24 hours.
    \item \textbf{dissatisfied energy gap (DEG):} Mean difference between target and achieved SoC values across all EVs.
    \item \textbf{average queue length (AQL):} Mean vehicle waiting queue length over a 24-hour period.
    \item \textbf{charging economic cost (ECC):} Total economic cost incurred by EVs for charging during 24 hours.
    \item \textbf{load pressure of the grid (LPG):} Grid load pressure during transmission, as defined in Eq.~(\ref{eqn:grid1}), measured over 24 hours.
\end{itemize}

Based on these five evaluation metrics representing different stakeholder perspectives, we present an additional comprehensive metric, the \textbf{charging market efficiency index (CMEI)}, which measures the overall operational efficiency of the EV charging market as follows:
\begin{equation}
\begin{aligned}
\text{CMEI} = &\frac{w_1(\text{DPS} - \text{DPS}_{min})}{\text{DPS}_{max} - \text{DPS}_{min}} + \frac{w_2(\text{DEG}_{max} - \text{DEG})}{\text{DEG}_{max} - \text{DEG}_{min}} \\
& + \frac{w_3(\text{AQL}_{max} - \text{AQL})}{\text{AQL}_{max} - \text{AQL}_{min}} + \frac{w_4(\text{ECC}_{max} - \text{ECC})}{\text{ECC}_{max} - \text{ECC}_{min}} \\
& + \frac{w_5(\text{LPG}_{max} - \text{LPG})}{\text{LPG}_{max} - \text{LPG}_{min}}
\end{aligned}
\end{equation}
where weights $w_1=w_5=1$ denote the coefficients for measuring market efficiency from the perspectives of the charging station and the grid, respectively, while $w_2=w_3=w_4=1/3$ denote the coefficients from the EV users' perspective.

\begin{figure}[!t]
\centering
\includegraphics[width=3.5in]{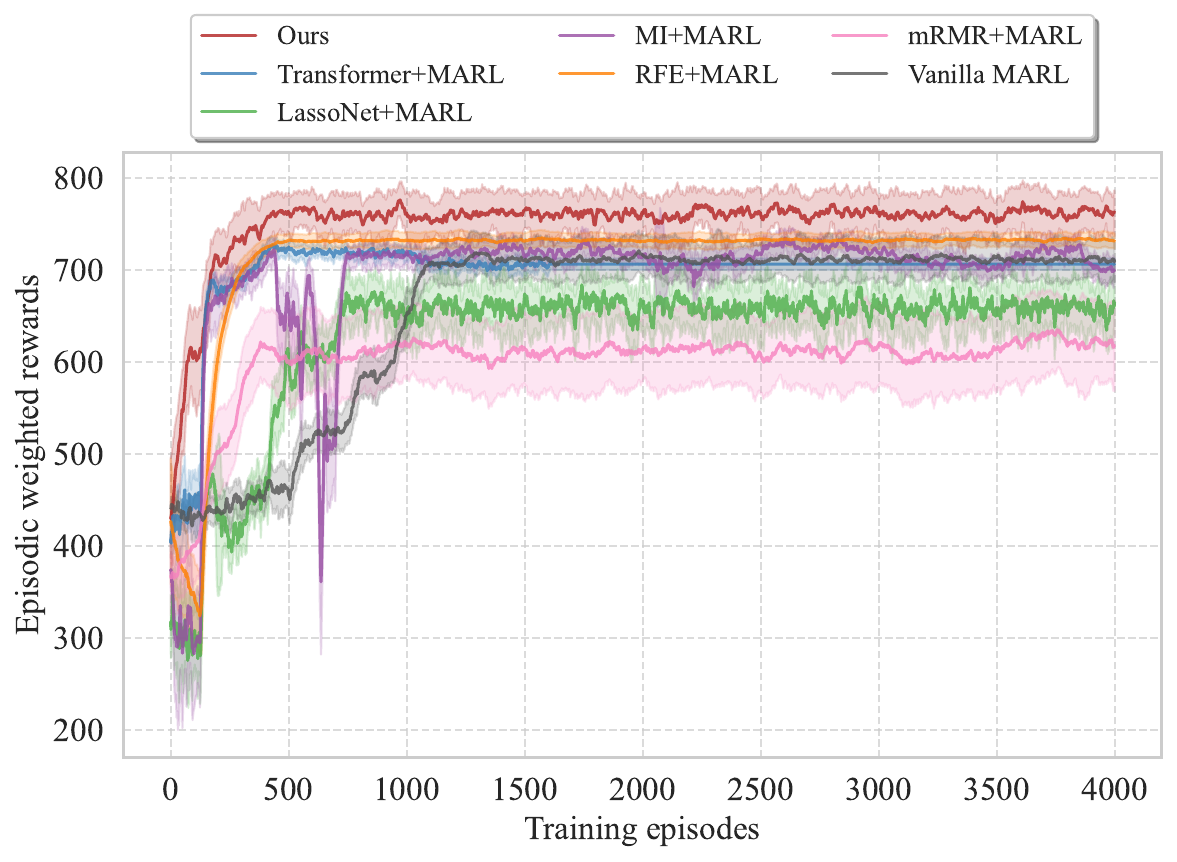}
\caption{Episodic weighted rewards during training for feature selection-based approaches with five random seeds.}
\label{fig:train_rewards_states}
\vspace{-4mm}
\end{figure}

\subsection{Comparison with Feature Selection-Based Approaches}

Figure~\ref{fig:train_rewards_states} shows the training process in terms of episodic weighted rewards (averaged over five random seeds) for our approach and several state-of-the-art feature selection-based methods. Our method converges after approximately 642 episodes, significantly faster than baselines such as MI+MARL and Vanilla MARL (which require over 1,200 episodes). These results demonstrate that our LLM-based feature selection effectively extracts the most relevant state information, facilitating more efficient learning.

\begin{table*}[!t]
\centering
\caption{Comparison of feature selection-based approaches evaluated using DPS, DEG, AQL, ECC, LPG, and CMEI metrics during the execution phase (averaged over five seeds). \textbf{Bold} values indicate the best results. The gray-highlighted column displays the CMEI, which measures the overall operational efficiency of the charging market.}
\label{tab:execution_rewards_states}
\begin{tabular}{lcccccc}
\toprule
Approach & DEG ($\downarrow$) & AQL ($\downarrow$) & ECC ($\downarrow$) & DPS ($\uparrow$) & LPG ($\downarrow$) & \cellcolor{gray!20} CMEI ($\uparrow$) \\
\midrule
%\rowcolor{gray!10} \multicolumn{7}{l}{\textbf{Proposed Approach}} \\
Ours (LLM-FS+LLM-MOB) & ${0.037 \pm 0.014}$ & $\mathbf{2.958 \pm 1.241}$ & $\mathbf{104.2 \pm 35.1}$ & ${86.9 \pm 40.2}$ & $\mathbf{0.411 \pm 0.021}$ & \cellcolor{gray!20} $\mathbf{2.195 \pm 0.192}$ \\
\hdashline
%\rowcolor{gray!10} \multicolumn{7}{l}{\textbf{LassoNet-based Methods}} \\
LassoNet+MARL & ${0.039 \pm 0.015}$ & ${3.479 \pm 1.435}$ & ${234.2 \pm 151.8}$ & ${13.6 \pm 17.0}$ & ${5.884 \pm 0.489}$ & \cellcolor{gray!20} ${1.705 \pm 0.316}$ \\
LassoNet+LLM-MOB & ${0.044 \pm 0.012}$ & ${5.083 \pm 1.782}$ & ${217.2 \pm 151.8}$ & ${28.7 \pm 15.6}$ & ${5.792 \pm 0.381}$ & \cellcolor{gray!20} ${1.672 \pm 0.217}$ \\
\hdashline
%\rowcolor{gray!10} \multicolumn{7}{l}{\textbf{RFE-based Methods}} \\
RFE+MARL & ${0.071 \pm 0.016}$ & ${6.987 \pm 0.523}$ & ${365.5 \pm 118.4}$ & ${138.3 \pm 45.2}$ & ${0.524 \pm 0.035}$ & \cellcolor{gray!20} ${1.896 \pm 0.185}$ \\
RFE+LLM-MOB & ${0.041 \pm 0.017}$ & ${3.979 \pm 2.628}$ & ${268.2 \pm 103.1}$ & ${42.1 \pm 162.7}$ & ${0.426 \pm 0.028}$ & \cellcolor{gray!20} ${2.002 \pm 0.372}$ \\
\hdashline
%\rowcolor{gray!10} \multicolumn{7}{l}{\textbf{Transformer-based Methods}} \\
Transformer+MARL & ${0.117 \pm 0.022}$ & ${10.251 \pm 0.923}$ & ${333.5 \pm 118.4}$ & $\mathbf{156.3 \pm 37.1}$ & ${0.416 \pm 0.035}$ & \cellcolor{gray!20} ${1.719 \pm 0.212}$ \\
Transformer+LLM-MOB & ${0.084 \pm 0.010}$ & ${7.271 \pm 1.081}$ & ${112.6 \pm 68.2}$ & ${42.7 \pm 22.4}$ & ${0.425 \pm 0.031}$ & \cellcolor{gray!20} ${1.869 \pm 0.127}$ \\
\hdashline
%\rowcolor{gray!10} \multicolumn{7}{l}{\textbf{MI-based Methods}} \\
MI+MARL & ${0.083 \pm 0.019}$ & ${7.625 \pm 0.622}$ & ${460.6 \pm 116.3}$ & ${82.1 \pm 48.7}$ & ${9.227 \pm 2.992}$ & \cellcolor{gray!20} ${1.238 \pm 0.353}$ \\
MI+LLM-MOB & $\mathbf{0.035 \pm 0.022}$ & ${4.111 \pm 2.462}$ & ${289.4 \pm 82.7}$ & ${138.1 \pm 124.2}$ & ${0.418\pm 0.031}$ & \cellcolor{gray!20} ${2.129 \pm 0.345}$ \\
\hdashline
%\rowcolor{gray!10} \multicolumn{7}{l}{\textbf{mRMR-based Methods}} \\
mRMR+MARL & ${0.044 \pm 0.006}$ & ${3.458 \pm 0.598}$ & ${321.0 \pm 93.2}$ & ${71.5 \pm 30.2}$ & ${9.971 \pm 3.257}$ & \cellcolor{gray!20} ${1.495 \pm 0.223}$ \\
mRMR+LLM-MOB & ${0.059 \pm 0.017}$ & ${5.677 \pm 1.967}$ & ${198.6 \pm 104.5}$ & ${137.1 \pm 24.5}$ & ${0.421 \pm 0.028}$ & \cellcolor{gray!20} ${2.065 \pm 0.194}$ \\
\hdashline
%\rowcolor{gray!10} \multicolumn{7}{l}{\textbf{Baseline}} \\
Vanilla MARL & ${0.089 \pm 0.038}$ & ${69.128 \pm 1.215}$ & ${251.4 \pm 40.6}$ & ${35.0 \pm 23.8}$ & ${0.746 \pm 0.043}$ & \cellcolor{gray!20} ${1.698 \pm 0.186}$ \\
\bottomrule
\end{tabular}
\vspace{-3mm}
\end{table*}

Table~\ref{tab:execution_rewards_states} details the execution-phase performance, measured via DPS, DEG, AQL, ECC, LPG, and CMEI. Our approach achieves superior AQL ($2.958 \pm 1.241$), ECC ($104.2 \pm 35.1$), and LPG ($0.411 \pm 0.021$). Although MI+LLM-MOB and Transformer+MARL yield marginally better DEG and DPS, our method attains the highest overall CMEI ($2.195 \pm 0.192$), indicating that our feature selection module successfully captures critical state information and further advances the market operation efficiency. Furthermore, comparing approaches with standard MARL against those enhanced by LLM-MOB reveals that the LLM-MOB integration commonly improves performance across different feature selection-based methods, demonstrating its effectiveness in achieving balanced multi-objective optimization.

\begin{table}[!t]
\centering
\caption{Average training efficiency metrics for feature selection-based approaches: episodes to convergence, time per episode (sec), and total training time (min). \textbf{Bold value} indicates the minimum total training time.}
\label{tab:time_state_methods}
\scalebox{1}{
\begin{tabular}{lccc}
\toprule
Approach & \makecell[c]{Episodes to\\converge} & \makecell[c]{Time per\\episode (sec)} & \makecell[c]{Total time\\(min)}\\ 
\midrule
%\rowcolor{gray!10} \multicolumn{4}{l}{\textbf{Proposed Approach}} \\
Ours & ${642}$ & ${3.85}$ & $\mathbf{41.16}$ \\
\hdashline
%\rowcolor{gray!10} \multicolumn{4}{l}{\textbf{LassoNet-based Approach}} \\
LassoNet+MARL & ${557}$ & ${7.63}$ & ${70.83}$  \\
LassoNet+LLM-MOB & ${536}$ & ${7.67}$ & ${68.52}$  \\
\hdashline
%\rowcolor{gray!10} \multicolumn{4}{l}{\textbf{RFE-based Approach}} \\
RFE+MARL & ${718}$ & ${5.04}$ & ${60.31}$ \\
RFE+LLM-MOB & ${581}$ & ${4.41}$ & ${42.70}$ \\
\hdashline
%\rowcolor{gray!10} \multicolumn{4}{l}{\textbf{Transformer-based Approach}} \\
Transformer+MARL & ${821}$ & ${40.41}$ & ${552.94}$ \\
Transformer+LLM-MOB & ${889}$ & ${35.37}$ & ${524.06}$ \\
\hdashline
%\rowcolor{gray!10} \multicolumn{4}{l}{\textbf{MI-based Approach}} \\
MI+MARL & ${754}$ & ${5.06}$ & ${63.59}$ \\
MI+LLM-MOB & ${731}$ & ${5.38}$ & ${65.55}$ \\
\hdashline
%\rowcolor{gray!10} \multicolumn{4}{l}{\textbf{MRMR-based Approach}} \\
mRMR+MARL & ${618}$ & ${8.36}$ & ${86.11}$ \\
mRMR+LLM-MOB & ${1170}$ & ${5.23}$ & ${101.98}$ \\
\hdashline
%\rowcolor{gray!10} \multicolumn{4}{l}{\textbf{Baseline}} \\
Vanilla MARL & ${1458}$ & ${6.19}$ & ${150.42}$   \\
\bottomrule
\end{tabular}}
\vspace{-3mm}
\end{table}

The computational efficiency comparison presented in Table~\ref{tab:time_state_methods} validates the superior performance of our approach. Our method achieves a remarkable training speed of 3.85 seconds per episode, substantially outperforming conventional feature selection-based methods. With rapid convergence in just 642 episodes, our approach attains the minimal total training time of 41.16 minutes. This computational advantage primarily stems from our novel design paradigm: unlike traditional feature selection methods that require repetitive feature extraction in each episode, our approach leverages semantic understanding and reasoning to perform feature selection only once prior to policy training.

%which performs neither feature selection nor multi-objective balancing

\subsection{Comparison with MORL-Based Approaches}

\begin{figure}[!t]
\centering
\includegraphics[width=3.5in]{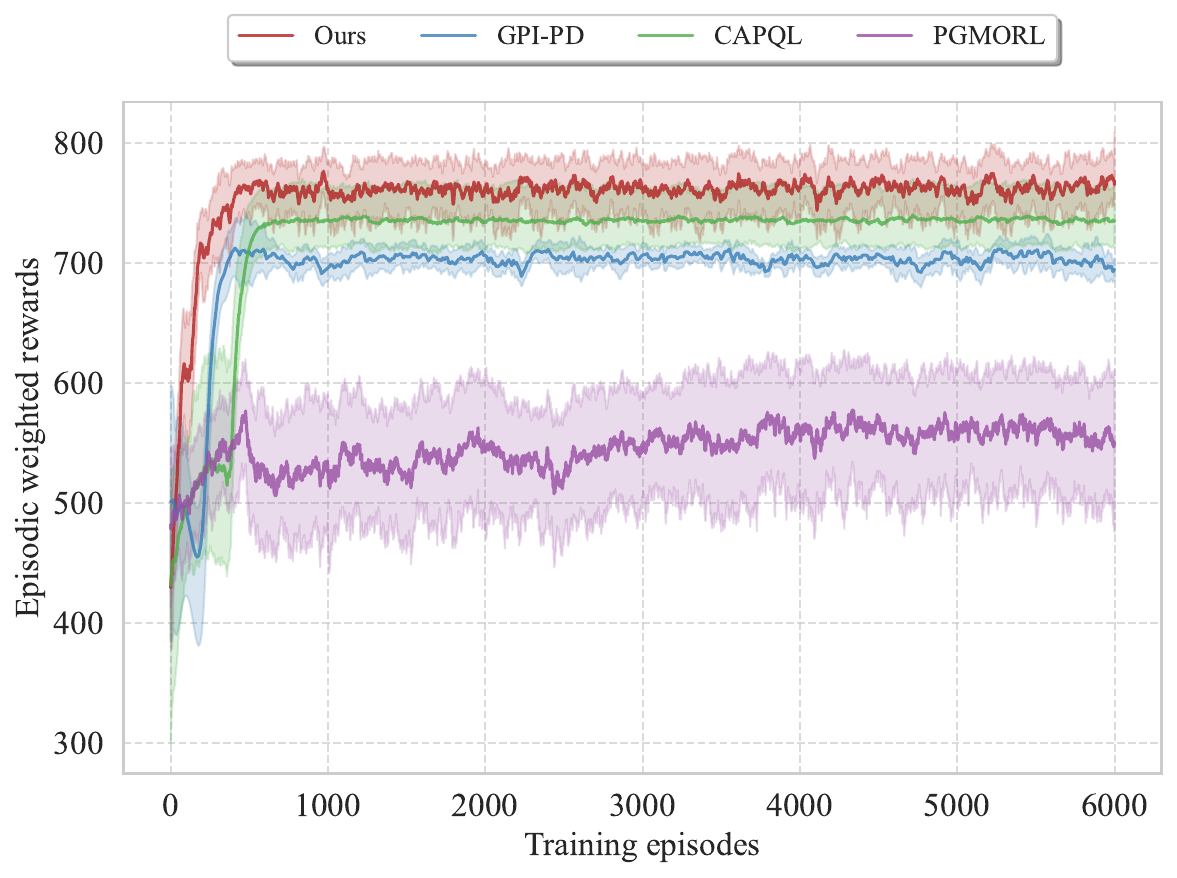}
\caption{Episodic weighted rewards during training for multi-objective RL-based methods over five random seeds.}
\label{fig:results_rewards_weights}
\vspace{-4mm}
\end{figure}

\begin{table*}[!t]
\centering
\caption{Comparison between our approach and MORL-based approaches evaluated using DPS, DEG, AQL, ECC, LPG, and CMEI metrics during execution (averaged over five random seeds).}
\label{tab:results_weights}
\scalebox{1}{
\begin{tabular}{lcccccc}
\toprule
Approach & DEG ($\downarrow$) & AQL ($\downarrow$) & ECC ($\downarrow$) & DPS ($\uparrow$) &  LPG ($\downarrow$) & \cellcolor{gray!20} CMEI ($\uparrow$) \\ % 将最后一列标题设置为灰色
\midrule
%\rowcolor{gray!10} \multicolumn{7}{l}{\textbf{Proposed Approach}} \\
Ours (LLM-FS+LLM-MOB) & ${0.037 \pm 0.014}$ & $\mathbf{2.958 \pm 1.241}$ & $\mathbf{104.2 \pm 35.1}$ & ${86.9 \pm 40.2}$ & $\mathbf{0.411 \pm 0.021}$ & \cellcolor{gray!20} $\mathbf{2.195 \pm 0.192}$\\
\hdashline
%\rowcolor{gray!10} \multicolumn{7}{l}{\textbf{GPI-PD-based Approach}} \\
GPI-PD & ${0.035 \pm 0.016}$ & ${2.983 \pm 0.791}$ & ${342.6 \pm 106.8}$  & ${4.6 \pm 6.1}$ & ${15.189 \pm 1.471}$ & \cellcolor{gray!20} ${1.153 \pm 0.274}$\\
GPI-PD+LLM-FS & $\mathbf{0.034 \pm 0.012}$ & ${4.104 \pm 1.385}$ & ${281.8 \pm 162.3}$  & ${56.7 \pm 12.5}$ & ${8.983 \pm 8.365}$ & \cellcolor{gray!20} ${1.555 \pm 0.480}$\\
\hdashline
%\rowcolor{gray!10} \multicolumn{7}{l}{\textbf{CAPQL-based Approach}} \\
CAPQL & ${0.072 \pm 0.023}$ & ${5.625 \pm 0.682}$ & ${146.7 \pm 60.2}$ & ${70.9 \pm 39.4}$  & ${0.425 \pm 0.041}$ & \cellcolor{gray!20} ${1.973 \pm 0.145}$ \\
CAPQL+LLM-FS & ${0.034 \pm 0.011}$ & ${4.333 \pm 0.973}$ & ${350.1 \pm 125.7}$ & $\mathbf{165.2 \pm 112.4}$  & ${2.525 \pm 1.941}$ & \cellcolor{gray!20} ${2.012 \pm 0.145}$ \\
\hdashline
%\rowcolor{gray!10} \multicolumn{7}{l}{\textbf{PGMORL-based Approach}} \\
PGMORL & ${0.047 \pm 0.014}$ & ${4.652 \pm 0.917}$ & ${303.4 \pm 41.7}$ & ${89.9 \pm 21.1}$ & ${5.012 \pm 1.368}$ & \cellcolor{gray!20} ${1.756 \pm 0.192}$ \\
PGMORL+LLM-FS & ${0.038 \pm 0.015}$ & ${4.010 \pm 1.217}$ & ${224.5 \pm 70.5}$ & ${31.6 \pm 10.8}$ & ${4.356 \pm 1.012}$ & \cellcolor{gray!20} ${1.812 \pm 0.192}$ \\
%\rowcolor{gray!10} \multicolumn{7}{l}{\textbf{Baseline}} \\
\hdashline
Vanilla MARL & ${0.089 \pm 0.038}$ & ${69.128 \pm 1.215}$ & ${251.4 \pm 40.6}$ & ${35.0 \pm 23.8}$ & ${0.746 \pm 0.043}$ & \cellcolor{gray!20} ${1.698 \pm 0.186}$ \\
\bottomrule
\end{tabular}}
\vspace{-4mm}
\end{table*}

Figure~\ref{fig:results_rewards_weights} compares the episodic weighted rewards of various MORL-based methods over five random seeds. Our approach exhibits more stable learning with higher reward levels compared to baseline methods. Specifically, CAPQL stabilizes at a lower reward level around 700, PGMORL shows high variance, and GPI-PD struggles to exceed 610 reward units. These results underscore our method’s superior ability to balance multiple competing objectives.

Table~\ref{tab:results_weights} presents the execution-phase performance evaluated using DPS, DEG, AQL, ECC, LPG, and CMEI. Although certain baseline methods excel in individual metrics (e.g., GPI-PD in DEG and AQL, CAPQL and PGMORL in DPS), they fail to optimize all objectives simultaneously. In contrast, our approach delivers a balanced performance across all key metrics, achieving the highest CMEI ($2.195 \pm 0.192$), outperforming the next best CAPQL+LLM-FS by 9.1\%. Moreover, comparing standard MORL-based methods with their variants enhanced by the LLM-FS module reveals that integrating our proposed feature selection technique significantly improves the performance of MORL approaches.

\begin{table}[!t]
\centering
\caption{Training efficiency metrics for multi-objective RL-based approaches (episodes to convergence, time per episode in seconds, and total training time in minutes).}
\label{tab:time_weights_methods}
\scalebox{1}{
\begin{tabular}{lccc}
\toprule
Approach & \makecell[c]{Episodes to\\converge} & \makecell[c]{Time per\\episode (sec)} & \makecell[c]{Total time\\(min)}\\ 
\midrule
%\rowcolor{gray!10} \multicolumn{4}{l}{\textbf{Proposed Approach}} \\
Ours & ${642}$ & ${3.85}$ & $\mathbf{41.16}$ \\
%\rowcolor{gray!10} \multicolumn{4}{l}{\textbf{GPI-PD-based Approach}} \\
\hdashline
GPI-PD & ${537}$ & ${16.40}$ & ${146.93}$  \\
GPI-PD+LLM-FS & ${626}$ & ${9.06}$ & ${94.53}$  \\
\hdashline
%\rowcolor{gray!10} \multicolumn{4}{l}{\textbf{CAPQL-based Approach}} \\
CAPQL & ${715}$ & ${8.95}$ & ${106.65}$ \\
CAPQL+LLM-FS & ${1328}$ & ${3.86}$ & ${85.43}$ \\
\hdashline
%\rowcolor{gray!10} \multicolumn{4}{l}{\textbf{PGMORL-based Approach}} \\
PGMORL & ${3885}$ & ${2.09}$ & ${135.33}$ \\
PGMORL+LLM-FS & ${3520}$ & ${0.82}$ & ${48.11}$ \\
\bottomrule
\end{tabular}}
\vspace{-2mm}
\end{table}

Table~\ref{tab:time_weights_methods} summarizes the training efficiency (episodes to convergence, time per episode, total training time) for our approach and MORL-based methods. Obviously, MORL-based approaches require significantly more episodes and longer training times than our proposed approach. Moreover, it can be observed that incorporating the LLM-FS module reduces both per-episode and overall training time, thus confirming its effectiveness in enhancing training efficiency.

\begin{figure}[!t]
\centering
\includegraphics[width=3in]{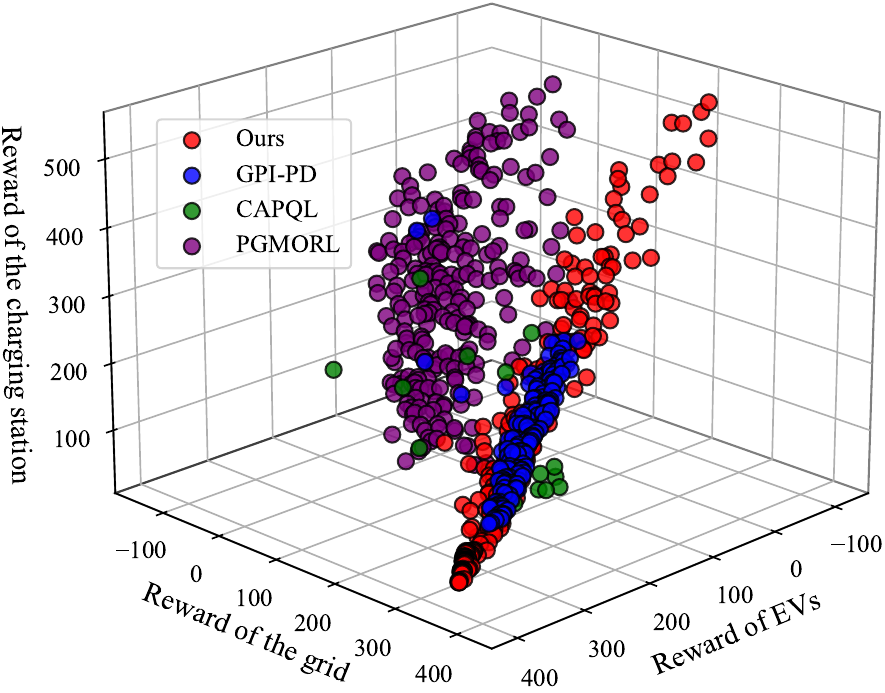}
\caption{Pareto fronts for different approaches across daily EVs, grid, and charging station rewards.}
\label{fig:results_pareto}
\vspace{-3mm}
\end{figure}

Figure~\ref{fig:results_pareto} illustrates the Pareto fronts of different approaches across three reward metrics: daily EVs reward, daily grid reward, and daily charging station reward. Our approach exhibits a more diverse Pareto front and consistently balances stakeholder rewards, i.e., delivering higher rewards for EVs and grid while maintaining competitive charging station performance. These advantages are attributable to: (1) the integration of expert knowledge via LLMs, which enhances both training and balancing efficiency; and (2) centralized multi-agent training that mitigates overfitting and increases Pareto diversity.

% \subsection{Visualisation Analysis}

% \subsubsection{}

\subsection{Interpretability Analysis}

A key advantage of our approach is the interpretability provided by the LLM-FS and LLM-MOB modules. We now present examples illustrating how the LLM explains its feature selection and multi-objective balancing decisions.

To elucidate the interpretability of our feature selection process, we present a case study demonstrating how LLM identifies and ranks the five most significant features from the EV perspective. Figure~\ref{fig:llm-feature-selection} illustrates this process, where \textbf{bold text} denotes the selected features and \textcolor{orange}{orange scores} indicate their relative importance. The accompanying rationale for each selection reveals the LLM's decision-making process, providing transparency into how the model evaluates features based on their physical significance and relevance to EV charging optimization.

\begin{figure}[htbp]
    \centering
    \begin{tcolorbox}[
        title={\textbf{LLM Feature Selection Output: EV Perspective}},
        colframe=blue!75!black,
        colback=blue!10,
        coltitle=white,
        fonttitle=\bfseries,
        fontupper=\small,
        boxrule=0.5mm,
        label=box:llm-feature-selection
    ]
    \begin{enumerate}[label=\textbf{\arabic*.}, leftmargin=*]
        \item \textbf{Desired SoC for charged EVs} \hfill \textcolor{orange}{(1.0)}\\
        \small{Critical as it directly influences the completion of the charging session and satisfaction of charging needs}
        
        \item \textbf{Desired departure time for charged EVs} \hfill \textcolor{orange}{(0.95)}\\
        \small{Very important for scheduling and minimizing waiting times and charging times}
        
        \item \textbf{Charging and waiting time cost weight factors} \hfill \textcolor{orange}{(0.9)}\\
        \small{These influence user satisfaction and optimization of time-based objectives}
        
        \item \textbf{Energy stored in the battery system of CS} \hfill \textcolor{orange}{(0.8)}\\
        \small{Determines available resources for charging and affects time reduction}
        
        \item \textbf{Charging efficiency for charged EVs} \hfill \textcolor{orange}{(0.75)}\\
        \small{Directly impacts the charging time and energy required}
    \end{enumerate}
    \end{tcolorbox}
    \caption{Feature importance ranking and rationale as determined by the LLM for EV charging scheduling.}
    \label{fig:llm-feature-selection}
\vspace{-4mm}
\end{figure}

\begin{figure}[htbp]
    \centering
    \begin{tcolorbox}[
        title={\textbf{LLM Multi-Objective Optimization Reasoning}},
        colframe=green!75!black,
        colback=green!5,
        coltitle=white,
        fonttitle=\bfseries,
        fontupper=\small,
        boxrule=0.5mm,
        label=box:llm-weight-decision
    ]
    \textbf{Reasoning:} With 18 EVs currently being charged, several have imminent departure times (one at 7 o'clock and several at 8 o'clock), indicating pressing needs to optimize charging power and minimize charging/waiting time. Since EV drivers are sensitive to charging times and costs, these should be prioritized. Given the variety of desired SOC levels and departure times, the focus should be on ensuring that EVs meet their energy requirements efficiently and affordably. The diverse coefficients signify urgency and variability in how charging time costs impact different vehicles, suggesting dynamic control of charging power to accommodate EV needs.
    
    \textbf{Decision:} weight \textcolor{orange}{1.5} for EVs standpoint.
    \end{tcolorbox}
    \caption{Weights balancing in multi-objective optimization as determined by the LLM for EV charging scheduling.}
    \label{fig:llm-weights}
\vspace{-3mm}
\end{figure}

Figure~\ref{fig:llm-weights} presents an example of the LLM’s reasoning for multi-objective weight assignment. Here, the LLM observes that 18 EVs are charging with several having imminent departure times (e.g., at 7 or 8 o'clock), which underscores an urgent need to optimize both charging power and waiting times. After evaluating driver sensitivity to charging duration and cost, the LLM assigns a weight of \textcolor{orange}{1.5} to the EV perspective.  This transparent reasoning chain provides insight into how the LLM balances competing objectives for different roles in EV charging market.

\subsection{Extension to Larger Scenarios}

To evaluate the scalability of our approach, we conduct extensive experiments with increasing numbers of EVs and charging stations. In multi-station scenarios, each EV $i \in \mathcal{V}$ determines its optimal charging station $cs^* \in \mathcal{CS}$ by minimizing a comprehensive priority index that incorporates three key factors: charging prices, queue length, and spatial distance:
\begin{equation}\label{eq:station_selection}
cs^* = \mathop {\arg \min }\limits_{cs \in \mathcal{CS}} \left\{ {\beta _1}p_{t,cs}^c + {\beta _2}{k_{t,cs}} + {\beta _3}D\left( {{l_{cs}},{l_i}} \right) \right\}, \forall i \in \mathcal{V}
\end{equation}
where $p_{t,cs}^c$ is the charging price at station $cs$ at time $t$, $k_{t,cs}$ denotes the queue length, $D\left( {{l_{cs}},{l_i}} \right)$ represents the Euclidean distance between EV location $l_i$ and station location $l_{cs}$, and $\beta_1$, $\beta_2$, $\beta_3$ are the weighting coefficients for these factors respectively.
For experimental setup, we randomly distribute both EVs and charging stations on a $100 \times 100$ grid map. To ensure balanced contributions from each factor in the priority index, we set $\beta_1 = 1$ as the baseline, while randomly sampling $\beta_2$ and $\beta_3$ from the intervals $[0,0.4]$ and $[0,0.04]$, respectively.

\begin{table*}[!t]
\centering
\caption{Performance Evaluation Across Different Scenario Scales}
\label{tab:scalability}
\scalebox{1}{
\begin{tabular}{llcccccc}
\toprule
\multicolumn{2}{c}{Scenario Scale} & \multicolumn{6}{c}{Performance Metrics} \\
\cmidrule(lr){1-2} \cmidrule(lr){3-8}
EV number & Station number & DEG ($\downarrow$) & AQL ($\downarrow$) & ECC ($\downarrow$) & DPS ($\uparrow$) & LPG ($\downarrow$) & \cellcolor{gray!20} CMEI ($\uparrow$) \\
\midrule
200+ & 1 & 0.037 ± 0.014 & 2.958 ± 1.241 & 104.2 ± 35.1 & 86.9 ± 40.2 & 0.411 ± 0.021 & \cellcolor{gray!20} 2.195 ± 0.192 \\
400+ & 1 & 0.228 ± 0.017 & 45.162 ± 2.226 & 137.6 ± 14.4 & 99.4 ± 15.2 & 0.487 ± 0.034 & \cellcolor{gray!20} 1.916 ± 0.155 \\
400+ & 3 & 0.052 ± 0.012 & 2.521 ± 1.951 & 529.2 ± 69.8 & 292.1 ± 142.1 & 3.791 ± 0.032 & \cellcolor{gray!20} 2.045 ± 0.454 \\
800+ & 3 & 0.162 ± 0.033 & 23.547 ± 5.346 & 545.6 ± 75.4 & 299.3 ± 164.6 & 4.093 ± 0.416 & \cellcolor{gray!20} 2.055 ± 0.441 \\
800+ & 5 & 0.067 ± 0.019 & 5.468 ± 0.832 & 1115.6 ± 314.1 & 751.1 ± 343.1 & 10.651 ± 0.128 & \cellcolor{gray!20} 2.114 ± 0.552 \\
1200+ & 5 & 0.116 ± 0.013 & 17.109 ± 1.995 & 1202.6 ± 279.8 & 809.5 ± 251.4 & 13.074 ± 0.761 & \cellcolor{gray!20} 2.206 ± 0.573 \\
\bottomrule
\end{tabular}}
\vspace{-3.5mm}
\end{table*}

Experimental results in Table~\ref{tab:scalability} reveal three key observations: (1) Our approach consistently maintains desired market operation efficiency (CMEI $>1.9$) across all tested scales; (2) When EV numbers increase while charging stations remain fixed, charging satisfaction metrics (DEG and AQL) significantly deteriorate due to intensified competition for limited resources. Meanwhile, station profits (DPS) and grid load pressure (ECC) exhibit only marginal growth, attributable to a limited increase in total charging volume; (3) Conversely, expanding station capacity while holding EV numbers constant substantially improves charging satisfaction (DEG and AQL) through increased charging resource availability, while simultaneously driving significant growth in both station profits (DPS) and grid load pressure (ECC).
 
\subsection{Ablation Studies}

\subsubsection{Component Analysis}

\begin{table*}[!t]
\centering
\caption{Ablation studies for component analysis.}
\label{tab:Component Analysis}
\scalebox{1}{
\begin{tabular}{lcccccc}
\toprule
Approaches & DEG ($\downarrow$) & AQL ($\downarrow$) & ECC ($\downarrow$) & DPS ($\uparrow$) &  LPG ($\downarrow$) & \cellcolor{gray!20} CMEI ($\uparrow$) \\ % 将最后一列标题设置为灰色
\midrule
Ours & $\mathbf{0.037 \pm 0.014}$ & $\mathbf{2.958 \pm 1.241}$ & $\mathbf{104.2 \pm 35.1}$ & ${86.9 \pm 40.2}$ & $\mathbf{0.411 \pm 0.021}$ & \cellcolor{gray!20} $\mathbf{2.195 \pm 0.192}$\\
w/o LLM-FS & ${0.059 \pm 0.006}$ & ${4.931 \pm 1.256}$ & ${230.3 \pm 105.2}$  & ${11.6 \pm 76.1}$ & ${6.428 \pm 1.467}$ & \cellcolor{gray!20} ${1.573 \pm 0.274}$\\
w/o LLM-MOB  & ${0.071 \pm 0.028}$ & ${7.275 \pm 4.142}$ & ${290.1 \pm 201.7}$ & $\mathbf{156.7 \pm 154.2}$  & ${3.941 \pm 3.500}$ & \cellcolor{gray!20} ${1.764 \pm 0.145}$ \\
\bottomrule
\end{tabular}}
\vspace{-3.5mm}
\end{table*}

To assess the contributions of our key modules, We compare our complete approach with two variants: \textbf{w/o LLM-FS:} The complete feature set is used instead of the LLM-selected subset; \textbf{w/o LLM-MOB:} Fixed weights are employed rather than adaptive balancing by LLM-MOB module. Results in Table~\ref{tab:Component Analysis}. show that our complete approach yields the best performance across five of six metrics (DEG, AQL, ECC, LPG, CMEI). Removing LLM-FS degrades metrics significantly, especially for LPG (rising from 0.411 to 6.428) and CMEI (dropping from 2.195 to 1.573). Although omitting LLM-MOB results in a higher DPS, other metrics (particularly AQL and ECC) suffer. These results verify that LLM-FS and LLM-MOB are both critical: LLM-FS enhances feature relevance and efficiency, while LLM-MOB enables balanced optimization.

\subsubsection{Performance with different LLMs}

\begin{table*}[!t]
\centering
\caption{Comparison of different LLMs employed in our approach evaluated using DPS, DEG, AQL, ECC, LPG, and CMEI metrics.}
\label{tab:different LLM}
\scalebox{1}{
\begin{tabular}{lcccccc}
\toprule
LLM & DEG ($\downarrow$) & AQL ($\downarrow$) & ECC ($\downarrow$) & DPS ($\uparrow$) &  LPG ($\downarrow$) & \cellcolor{gray!20} CMEI ($\uparrow$) \\ % 将最后一列标题设置为灰色
\midrule
GPT-4o & $\mathbf{0.037 \pm 0.014}$ & $\mathbf{2.958 \pm 1.241}$ & ${104.2 \pm 35.1}$ & ${86.9 \pm 40.2}$ & ${0.411 \pm 0.021}$ & \cellcolor{gray!20} ${2.195 \pm 0.192}$\\
o3-mini & ${0.060 \pm 0.016}$ & ${4.291 \pm 0.582}$ & ${349.8 \pm 43.6}$  & $\mathbf{279.4 \pm 48.9}$ & ${0.432 \pm 0.004}$ & \cellcolor{gray!20} $\mathbf{2.215 \pm 0.155}$\\
DeepSeek-R1 & ${0.065 \pm 0.005}$ & ${6.625 \pm 2.215}$ & ${115.9 \pm 43.2}$ & ${68.23 \pm 27.3}$  & ${0.411 \pm 0.006}$ & \cellcolor{gray!20} ${1.984 \pm 0.148}$ \\
LLaMA3-70B & ${0.066 \pm 0.019}$ & ${6.000 \pm 1.875}$ & ${272.9 \pm 170.5}$ & ${199.9 \pm 21.1}$ & $\mathbf{0.410 \pm 0.007}$ & \cellcolor{gray!20} ${2.081 \pm 0.205}$ \\
Gemini-1.5 & ${0.087 \pm 0.018}$ & ${8.395 \pm 1.112}$ & $\mathbf{90.3 \pm 18.5}$ & ${32.56 \pm 10.3}$ & ${0.413 \pm 0.010}$ & \cellcolor{gray!20} ${1.824 \pm 0.082}$ \\
\bottomrule
\end{tabular}}
\vspace{-3.5mm}
\end{table*}

Table~\ref{tab:different LLM} compares the performance of our approach when using different LLMs for feature selection and multi-objective balancing. All tested LLMs achieve satisfactory overall performance (CMEI between 1.82 and 2.22). Specifically, GPT-4o achieves the best DEG (0.037) and AQL (2.958), while o3-mini attains the highest DPS (279.4) and CMEI (2.215). Gemini-1.5 demonstrates the lowest ECC (90.3), and LLaMA3-70B yields the best LPG (0.410). These results indicate that, while specific metrics vary with the choice of LLM, our approach remains robust across different LLM implementations. Notably, newer models (e.g., o3-mini released on January 31, 2025 by OpenAI~\cite{openaiOpenAIO3mini}) show promise in enhancing overall market efficiency despite somewhat higher computational costs.

\section{Conclusion}\label{sec:con}

In this paper, we proposed an LLM-enhanced multi-agnet reinforcement learning approach for unified EV charging scheduling in public charging systems. Our framework first simultaneously considers three key stakeholders: EVs, the charging station, and the power grid, as well as introduces two major innovations: a LLM-FS module for efficient extraction of relevant features in large-scale state spaces, and a LLM-MOB module for adaptive balancing among competing stakeholder objectives. Leveraging expert knowledge embedded in LLMs, our approach not only enhances the training efficiency of MARL algorithms but also offers interpretable solutions through natural language explanations. Extensive experimental results demonstrate that our method outperforms various state-of-the-art feature selection-based and multi-objective optimization-based approaches across multiple performance metrics.

%While our current framework can be straightforwardly extended to multi-station scenarios by treating each station as an independent entity with local optimization, future work will focus on developing more sophisticated charging scheduling strategies among multiple stations. This includes investigating optimal spatial resource allocation for EVs and pricing competition for stations to better serve the practical needs of urban transportation systems.

\bibliographystyle{IEEEtran}
\bibliography{refs}

\end{document}